\documentclass[runningheads]{llncs}
\usepackage{graphicx}

\usepackage[dvipsnames]{xcolor}
\usepackage{tikz}
\usepackage{comment}
\usepackage{amsmath,amssymb} 
\usepackage{color}
\usepackage{microtype}
\usepackage{graphicx}
\usepackage{subfigure}
\usepackage{rotating}
\usepackage{hyperref}

\usepackage{amsmath,amsfonts,bm}

\def\eqref#1{equation~\ref{#1}}

\def\1{\bm{1}}

\newcommand{\test}{\mathcal{D_{\mathrm{test}}}}

\DeclareMathAlphabet{\mathsfit}{\encodingdefault}{\sfdefault}{m}{sl}
\SetMathAlphabet{\mathsfit}{bold}{\encodingdefault}{\sfdefault}{bx}{n}

\DeclareMathOperator*{\argmin}{arg\,min}

\usepackage{caption}
\usepackage{lipsum}
\usepackage{mathabx}
\usepackage{array}
\usepackage{amsmath}
\usepackage{multirow}
\usepackage{booktabs} 
\usepackage{amsmath,graphicx,xcolor,booktabs,amsmath, graphbox}
\usepackage{soul}

\definecolor{Highlight}{HTML}{39b54a}  

\usepackage[accsupp]{axessibility}  

\newcommand{\norm}[1]{\left\lVert#1\right\rVert}
\begin{document}
\pagestyle{headings}
\mainmatter
\def\ECCVSubNumber{1135}  

\title{Image-based CLIP-Guided Essence Transfer} 

\titlerunning{Image-based CLIP-Guided Essence Transfer}
%
\author{Hila Chefer\inst{1} \and
Sagie Benaim\inst{2} \and
Roni Paiss\inst{1}
\and
Lior Wolf\inst{1}}
\authorrunning{H. Chefer et al.}
%
\institute{Tel Aviv University
\and
University of Copenhagen
}
\maketitle

\begin{figure*}[]
\begin{center}
\begin{tabular}{c@{~}c@{~}c@{~}c@{~}c@{~}c@{~}c@{~}c}
{\begin{turn}{90} ~~Target \end{turn}}&
\includegraphics[width=0.11\linewidth] {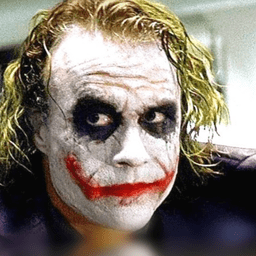}&
\includegraphics[width=0.11\linewidth] {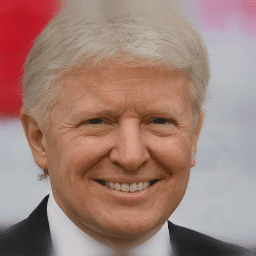}&
\includegraphics[width=0.11\linewidth] {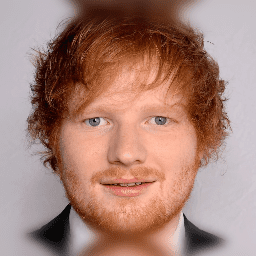}&
\includegraphics[width=0.11\linewidth] {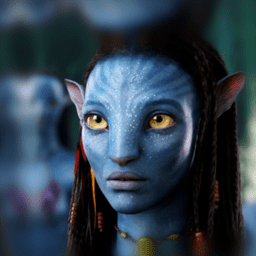}&
\includegraphics[width=0.11\linewidth] {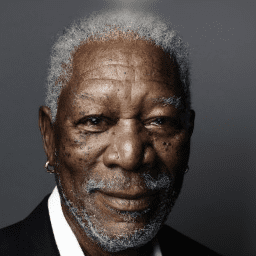}&
\includegraphics[width=0.11\linewidth] {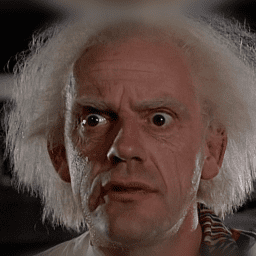}&
\includegraphics[width=0.11\linewidth] {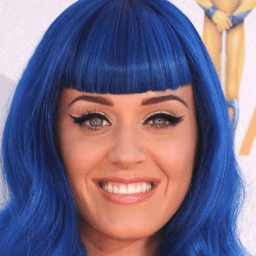}\\
{\begin{turn}{90} ~~Source \end{turn}}&
\includegraphics[width=0.11\linewidth] {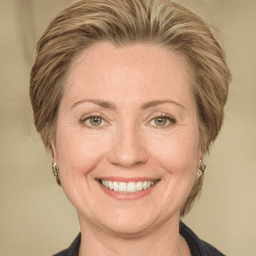}&
\includegraphics[width=0.11\linewidth]  {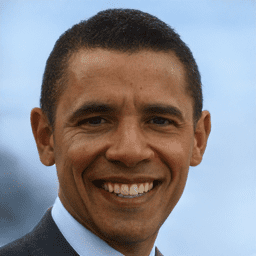} &
\includegraphics[width=0.11\linewidth] {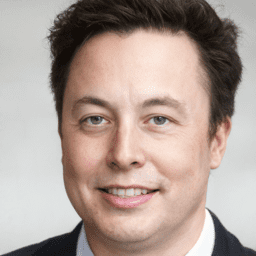} &
\includegraphics[width=0.11\linewidth] {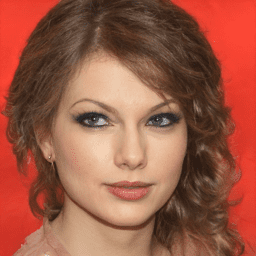} &
\includegraphics[width=0.11\linewidth] {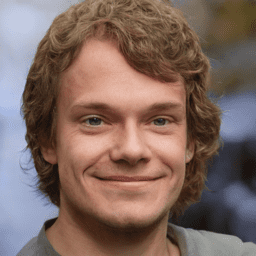}&
\includegraphics[width=0.11\linewidth] {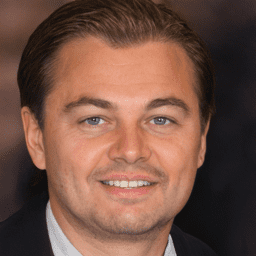}&
\includegraphics[width=0.11\linewidth] {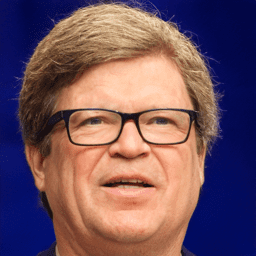}\\
{\begin{turn}{90} ~~Result  \end{turn}} &
\includegraphics[width=0.11\linewidth] {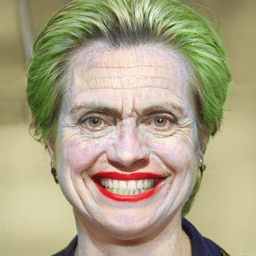} &
\includegraphics[width=0.11\linewidth]{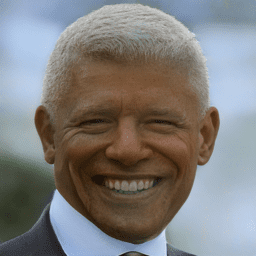} &
\includegraphics[width=0.11\linewidth]{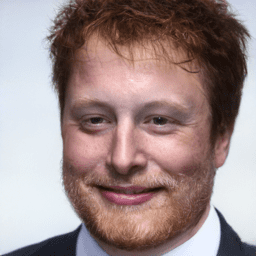} &
\includegraphics[width=0.11\linewidth]{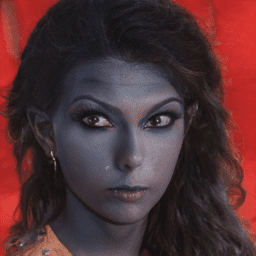} &
\includegraphics[width=0.11\linewidth] {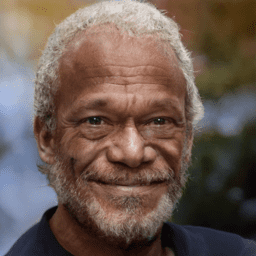} &
\includegraphics[width=0.11\linewidth] {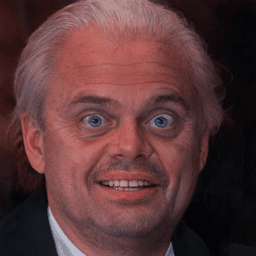}&
\includegraphics[width=0.11\linewidth] {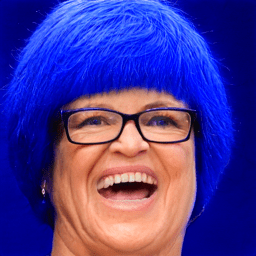}
\end{tabular}
    \caption{Results of our method on various targets and sources. The first row presents the target images to extract the essence from, the middle row shows the sources to transfer the essence to, and bottom row presents the results of our method.} 
    \label{fig:fig1}
\end{center}
\end{figure*}

\begin{abstract}
We make the distinction between (i) style transfer, in which a source image is manipulated to match the textures and colors of a target image, and (ii) essence transfer, in which one edits the source image to include high-level semantic attributes from the target. Crucially, the semantic attributes that constitute the essence of an image may differ from image to image. Our blending operator combines the powerful StyleGAN generator and the semantic encoder of CLIP in a novel way that is simultaneously additive in both latent spaces, resulting in a mechanism that guarantees both identity preservation and high-level feature transfer without relying on a facial recognition network. We present two variants of our method. The first is based on optimization, while the second fine-tunes an existing inversion encoder to perform essence extraction. Through extensive experiments, we demonstrate the superiority of our methods for essence transfer over existing methods for style transfer, domain adaptation, and text-based semantic editing. Our code is available at: https://github.com/hila-chefer/TargetCLIP.
\end{abstract}

\section{Introduction}
\label{sec:intro}
Style transfer, which typically refers to rendering the content of an image in the style of a different image, is a highly researched task in computer vision and computer graphics~\cite{Gatys_2016_CVPR,Bousmalis2017UnsupervisedPD,Ojha2021FewshotIG,Huang2017ArbitraryST,Efros2000,Hertzmann2000,li2017universal,luan2017deep,luan2018deep,sunkavalli2010multi,Chong2021JoJoGANOS}. This work explores a related task, which we refer to as \emph{essence transfer}. The essence of an image is defined to be the set of attributes that appear in the high-level textual description of the image. Our blending involves borrowing semantic features from a ``target'' image $I_t$ and transferring them to a ``source'' image $I_s$, thus creating an output image $I_{s,t}$. We find that ``essence" features capture properties such as skin complexion or texture, as do traditional style transfer methods, but also  semantic elements such as gender, age, and unique facial attributes, when considering faces. 

A rigorous definition of our goal is elusive, as the set of features defined as the essence may change from image to image, so we adopt a pragmatic approach. It has been shown~\cite{patashnik2021styleclip,gal2021stylegannada} that latent spaces of high-level vision networks,  (i.e., involving capabilities such as image understanding~\cite{ullman2000high}) are additive. Ergo, subtracting the representation of two inputs yields a meaningful shift between the inputs encoding the difference between the two. Our method transforms source images $I_s$ conditioned on a target image $I_t$. It forces the learned transformation to be doubly additive: once in the latent space of the generator, and once in the latent space of the understanding engine. Out of all possible ways of transforming a source image $I_s$ according to a target image $I_t$, we obtain, for every $I_t$ a transformation that is based on a constant shift over all $I_s$ in the generator space and leads to a constant difference in the high-level description of the image. 

For the generator network, we employ the powerful StyleGAN~\cite{karras2020analyzing} generator. Additivity in the latent space of StyleGAN was demonstrated in~\cite{shen2021closed,voynov2020unsupervised,harkonen2020ganspace}, for linearly interpolating  different images along semantic directions, as well as for the manipulation of semantic attributes (\cite{patashnik2021styleclip} for example). For the image recognition engine, we employ the CLIP network~\cite{radford2021learning}, which has shown impressive zero-shot capabilities across multiple domains such as image classification~\cite{radford2021learning} and adaptation of generated images~\cite{gal2021stylegannada}. It was also shown to behave additively~\cite{patashnik2021styleclip,gal2021stylegannada}. Since CLIP was trained in a contrastive manner, using textual descriptions, different images with the same high-level textual description are expected to receive a high similarity score, as their textual descriptions will be nearly identical. This allows our method to enforce consistency based on the semantic properties of the image, rather than pixel-level similarity.

We propose a method based on two loss terms. The first term ensures that the transformed image is semantically similar to the target image $I_t$ in the latent space of CLIP. The second term links the constant shift in the latent space of the generator to a constant shift in the latent space of CLIP, leading to a semantically consistent edit that is independent of the identity of the source $I_s$. 
We propose two methods for essence transfer. The first is based on per target optimization, while the second fine-tunes an inversion encoder to perform essence transfer. While an optimization-based approach is more accurate in capturing the relevant semantic properties, our encoder version only requires a forward pass on the target image to produce the target-specific (source-agnostic) essence vector, which defines the essence transfer operator for that image. 
We compare our method with state of the art style transfer and semantic editing methods and show that our novel double-additive formulation is necessary to successfully perform essence transfer. 
Finally, we decode to text the learned directions, demonstrating that the semantic edits we employ correspond to the attributes of the target image.

\section{Related Work}

\noindent{\it Style Transfer\quad}
Our work most closely relates to style transfer~\cite{Efros2000,Hertzmann2000,li2017universal,luan2017deep,sunkavalli2010multi,luan2018deep,Kim20DST}. 
Unlike \cite{Gatys_2016_CVPR,johnson2016perceptual,adain}, we derive style from CLIP~\cite{radford2021learning}, a recently proposed method for the semantic association of text and images. In this space, two images are close, or similar, to each other if their textual association is close. Such similarity may consider unique style elements, such as texture or complexion. It may also consider semantic elements, such as gender and facial attributes. We argue that this notion of style, which we use here, is more general. CLIP has been used in several recent works to enable the fine-tuning of StyleGAN for domain adaptation~\cite{gal2021stylegannada,Zhu2021MindTG} with impressive results, yet, as we show, the existing style transfer and domain adaptation methods fall short on the task of essence transfer, focusing usually on colors, textures and domain shifts, and can suffer from severe identity loss. 

\noindent{\it Image Manipulation\quad}
Our work is also related to recent image manipulation works based on a pre-trained generator~\cite{collins2020editing,tewari2020stylerig,harkonen2020ganspace,patashnik2021styleclip}. One set of works typically manipulates an image based on finding a set of possibly disentangled and semantic directions~\cite{shen2021closed,voynov2020unsupervised}. These works typically borrow the semantic meaning from the generator itself. A recent work called StyleCLIP~\cite{patashnik2021styleclip} showed the remarkable ability to borrow the semantic meaning from the CLIP space, and inspired many additional works to use CLIP for semantic editing and domain adaptation~\cite{Zhu2021MindTG,gal2021stylegannada,Abdal2021CLIP2StyleGANUE}. Our method uses CLIP in a similar manner to that presented in ~\cite{patashnik2021styleclip}. However, unlike our method, StyleCLIP considers text-driven manipulations, thus it is limited by what can be described in words, and the knowledge obtained by CLIP.  

\noindent{\it GAN Inversion\quad}
GAN inversion aims to extract a latent vector $z$ that corresponds to a target image $I$, i.e. $z$ holds that $G(z)=I$, where $G$ is the generator. Most inversion methods can be split into two types; optimization-based methods~\cite{Zhu2016GenerativeVM,Abdal2019Image2StyleGANHT,Bau2019SemanticPM,Creswell2019InvertingTG,Gu2020ImagePU,Lipton2017PreciseRO,Zhu2020ImprovedSE}, which employ an optimization process to find a latent $z$ such that $G(z)$ is closest to a specific target image $I$, and encoders~\cite{tov2021designing,Alaluf2021ReStyleAR,Guan2020CollaborativeLF,Kang2021GANIF,Kim2021ExploitingSD,luo2017learning,Perarnau2016InvertibleCG,Pidhorskyi2020AdversarialLA,Wang2021HighFidelityGI} which are trained to extract a latent $z$ for any input image $I$. Most methods for StyleGAN inversion focus on the $\mathcal{W}, \mathcal{W+}$ latent spaces. The $\mathcal{W}$ space is more editable, yet suffers from degraded expressiveness~\cite{Abdal2019Image2StyleGANHT}, therefore $\mathcal{W}+$ has been adopted for inversion. We employ the e4e encoder~\cite{tov2021designing} since it mitigates the distortion-editability trade-off by training an encoder in the $\mathcal{W}+$ latent space while encouraging the result to be close to the $\mathcal{W}$ space.

\section{Assumptions and Problem Formulation}
\label{sec:problem}
We now provide a formal definition of the essence transfer task, and an overview of the proposed method for any generator $G$ and semantic encoder $C$.

We define the essence of an image $I$ as the set of semantic features that constitute the high-level textual description of the image. The method employs four input components: (i) A generator $G$, which, given a vector $z$, generates an image $G(z)$, (ii) An image recognition engine $C$, which, given an image $I$, provides a latent representation of its high-level textual description, $C(I)$, in some latent space, (iii) A target image $I_t$, from which the essence is extracted, and (iv) A set of source images $S$, which are used to provide the statistics of images for which the method is applied. 
For clarity, we define $S$ as a set of $z$ vectors in a latent space of $G$, and each source image as $G(z)$.

Given these four inputs, our goal is to provide a generator $H$ such that the image $H(z)$ transfers the high-level textual features of a target image $I_t$ to the source image $I_s = G(z)$. 
If one wishes to transform an image $I_s$ rather than a vector $z$ using $H$, the image can be converted to a latent $z$ using an inversion method~\cite{tov2021designing,richardson2021encoding,roich2021pivotal}. We note that our formulation does not require, at any stage, the inversion of $I_t$.
On the generator side, linearity is expressed by:
\begin{equation}
\label{eq:H}
    H(z) = G(z+b)
\end{equation}
for some shift vector $b$, in the latent space of $G$.
Linearity in the latent space of the image recognition engine is expressed as:
\begin{equation}
  \forall z\in S \quad  d = C(H(z))-C(G(z))\,, 
  \label{eq:d}
\end{equation}
for some fixed $d$. Put differently, modifying any vector $z$ in $G$'s latent space with $b$ induces a constant semantic change in the latent space of $C$. This property is what is referred to as a ``global semantic direction'' by Patashnik et al.~\cite{patashnik2021styleclip}. However, our method goes about obtaining said global direction differently.  
The source-agnostic behavior is obtained by minimizing the following over $H$ and $d$:
\begin{equation}
  \sum_{z\in S} dist(C(H(z))-C(G(z)),d)\,,
\end{equation}
where $dist(\cdot,\cdot)$ calculates the distance between two vectors in the latent space of the semantic encoder $C$. For example, CLIP uses cosine similarity to estimate vector similarities.

Equivalently, to obtain $H$, we can minimize the following over $H$:
\begin{equation}
\label{eq:pairedconst}
  \sum_{z1,z2\in S} dist\left(C(H(z_1))-C(G(z_1)),C(H(z_2))-C(G(z_2))\right)\,. 
\end{equation}
So far, we defined the problem of learning a pair of semantic directions $b,d$, in two different latent spaces, such that $(b,d)$ match. We wish to add a constraint that ties the shifts to $I_t$. To this end, we wish to maximize similarity in the semantic space provided by the recognition engine $C$ between $I_t$ and the generated images $H(z)$. That is, we wish to minimize the sum of distances $\sum_{z \in S} dist(C(H(z)),C(I_t)) $.

\section{Method}
\begin{figure}[t!]
    \begin{center}
    \renewcommand{\arraystretch}{0.2}
\includegraphics[width=1\linewidth,clip] {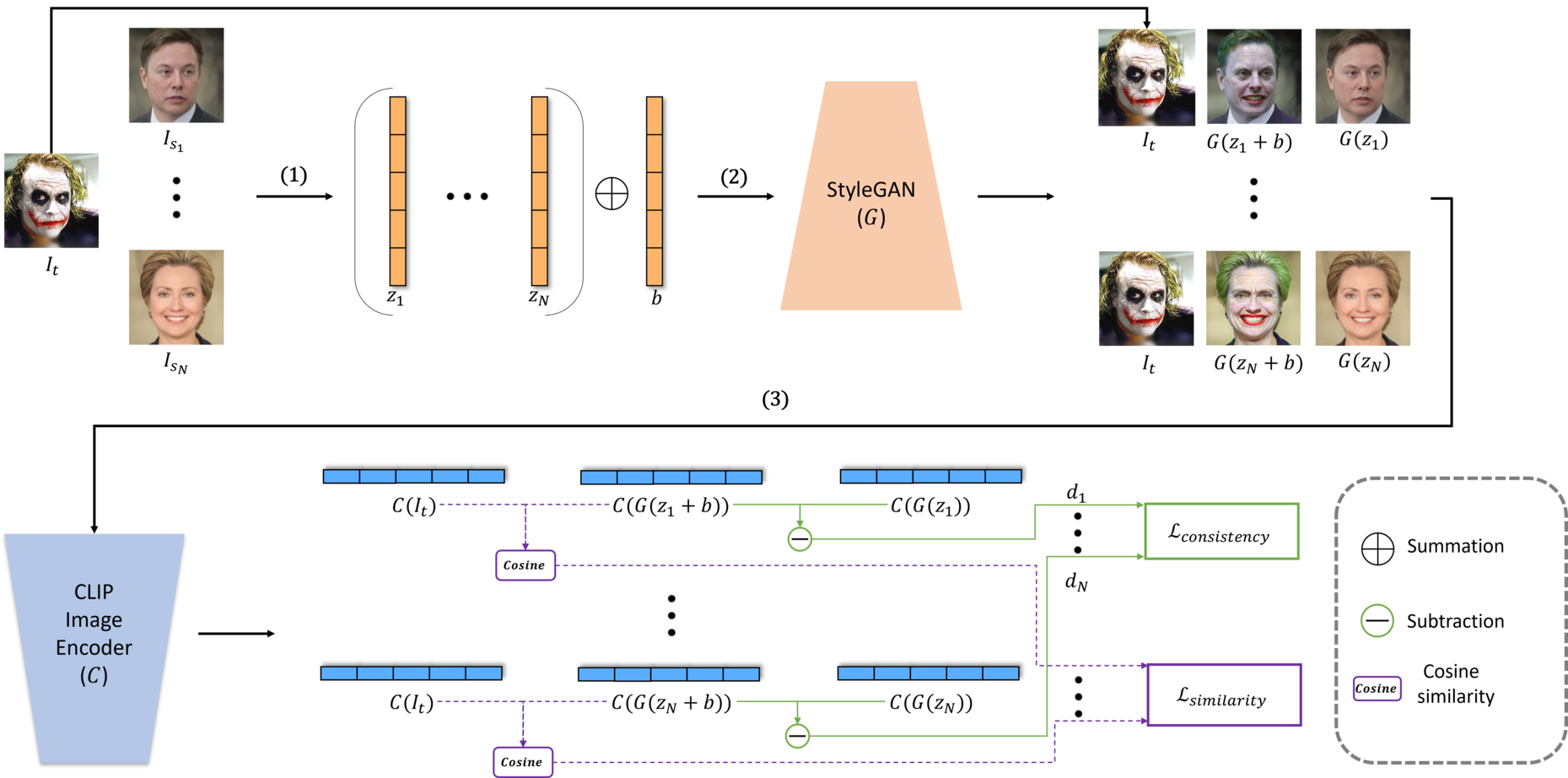}
    \captionof{figure}{An illustration of our loss calculation flow. Step (1) inverts the source images to obtain their latents $z_1,...,z_N$, and adds the proposed essence vector $b$. In step (2), the StyleGAN generator decodes the source latents $z_1,...,z_N$ and the manipulations $z_1+b,...,z_N+b$ to images. Step (3) encodes the sources ($G(z_1),...,G(z_N)$), manipulations ($G(z_1+b),..,G(z_N+b)$), and target ($I_t$) with CLIP. $\mathcal{L}_{consistency}$ demands that semantic changes be identical for $z_1,..,z_N$. $\mathcal{L}_{similarity}$ ensures that  $G(z_1+b),...,G(z_N+b)$ are semantically similar to $I_t$.
    }
    \label{fig:losses}
\end{center}
    \end{figure}

We now define our method, based on the formulation in Sec.~\ref{sec:problem}. Note that in accordance with the training process of CLIP, we employ cosine similarity to estimate semantic similarity between two images. 
$\sum_{z \in S} dist(C(H(z)),C(I_t)) $ becomes the following loss term, applied over a batch of source images, $S$:
\begin{align}
    \mathcal{L}_{similarity}=
    \frac{1}{N} \left(\sum_{z\in S} 1-  \frac{C(I_t)\cdot C(G(z + b))}{\norm{C(I_t)}_2\norm{C(G(z+b))}_2}\right),
    \label{eq:transfer-loss}
\end{align}
where $N=|S|$ is the batch size, and $z_1,..,z_N\in \mathcal{W+}$. The similarity loss estimates the semantic similarity between the image encodings of the target image and the manipulated images. By setting $N=1$, this loss becomes identical to the semantic loss employed by other semantic editing methods based on CLIP~\cite{patashnik2021styleclip}. 

The second concept we enforce is consistency. The goal of essence transfer is to modify the source image using a collection of semantic attributes that encapsulates the essence of the target image. These attributes are independent of the source image. We demand that the semantic edits induced by the direction $b$ in the latent space of $G$ be consistent across the source images, using CLIP's latent space. This is expressed in Eq.~\ref{eq:pairedconst} above and translates to the following loss:
\begin{align}
    \label{eq:consistency-loss}
      \mathcal{L}_{consistency} = 
     \frac{1}{{N\choose 2}} \left(\sum_{i_{src_1}, i_{src_2} \in I_s} 1- \frac {\Delta i_{src_1} \cdot \Delta i_{src_2}}  {\norm{\Delta i_{src_1}}_2\norm{\Delta i_{src_2}}_2}\right)
\end{align}
where $\Delta i_{src} = C(G( i_{src} +b))-C(G( i_{src}))$, as annotated in Eq.~\ref{eq:d} as $d$, and, as before, $N$ is the batch size. $\mathcal{L}_{consistency}$ guarantees that the direction encapsulated in $b$ produces semantic edits that are identical across a batch of source images $S$. 
Fig.~\ref{fig:losses} illustrates the steps of obtaining $\mathcal{L}_{similarity}, \mathcal{L}_{consistency}$ from a batch of sources $I_{s_1},...,I_{s_N}$ and a target image $I_t$.

The optimization problem solved during training in order to recover $H$, as defined in Eq.~\ref{eq:H}, is given as:
\begin{equation}
    b^{*} = \argmin \mathcal{L}_{similarity} +  \lambda_{consistency} \mathcal{L}_{consistency} + \lambda_{L_2}\norm{b}_2,
    \label{eq:objective}
\end{equation}
where $\lambda_{consistency}, \lambda_{L_2}$ are hyperparameters. We use the same hyperparameter values in all our experiments and all our methods: $\lambda_{consistency}=0.5, \lambda_{L_2}=0.003$. 

In contrast to other methods~\cite{patashnik2021styleclip}, ours does not rely on any face recognition models for preventing identity loss. In order to maintain the identity of the source images $I_1,...,I_N$ we employ a standard $L_2$ regularization to limit the magnitude of the effect that $b$ has on source images.

Restating Eq.~\ref{eq:H}, after obtaining the essence vector $b^*$ for a target image $I_t$, manipulating a source image $I_s$ is done as follows:
\begin{equation}
    I_{s,t} = G(z_s + b^*),
    \label{eq:manipulation}
\end{equation}
where $z_s$ is the latent that corresponds to the source image $I_s$, which can be obtained by inverting the image $I_s$. Thus, we simply add the essence vector $b^*$ to the latent representing the source image.

\noindent{\it Essence Optimization\quad} The first method we propose is a simple optimization process of finding an essence vector $b^*$ that minimizes the objective in Eq.~\ref{eq:objective}. Unlike other optimization-based methods for semantic editing, our method is more stable in the sense that the same set of hyperparameters can be applied for each target, and no target-specific tuning is required. The implementation employs the Adam optimizer~\cite{Kingma2015AdamAM} for $1000$ iterations with a learning rate of $0.2$. Due to resource limitations, we use only $N=4$ images for our double additivity losses (Eq.~\ref{eq:transfer-loss},~\ref{eq:consistency-loss}). For difficult edits, i.e. edits containing unconventional or extreme semantic attributes such as blue skin, we found it beneficial to initialize the direction $b$ in the optimization process to be the inversion of the target produced with the e4e encoder. This can be attributed intuitively to the fact that the inversion of the target contains, among other identity-specific attributes, the semantic attributes that constitute the high-level textual description of the image, i.e. its essence. Therefore, initializing the direction $b$ to be the inversion of the target steers the optimization toward semantic properties that are related to the target image. 

\noindent{\it Essence Encoder\quad}
For our second method, we fine-tune a pre-trained e4e encoder~\cite{tov2021designing} over the pSp framework~\cite{richardson2021encoding} to output the essence vector $b^*$ of the input image instead of its inversion. 
Since the encoder is pre-trained for inversion, the initial output for each image $I_t$ contains, among other features, the semantic features that comprise its essence. The goal of the fine-tuning process is to shift the weights of the encoder such that the output for each image $I_t$ will be the semantic parts of the inversion that correspond to the essence vector. 
This fine-tuning is performed on a small dataset of $200$ random images from the CelebA-HQ dataset~\cite{liu2015faceattributes}, and evaluated on $50$ random images from the CelebA-HQ test set. We use a learning rate of $1e-4$ for $3000$ iterations, with a batch size of $1$ target image and $N=5$ source images for our double additivity losses (Eq.~\ref{eq:transfer-loss},~\ref{eq:consistency-loss}). The objective and its hyperparameters are identical to the ones used for the optimization-based method (Eq.~\ref{eq:objective}). 
Unlike other methods that train an encoder or a generator for each target text or image, such as~\cite{patashnik2021styleclip,gal2021stylegannada}, our encoder is fine-tuned once and can accommodate \emph{any} target after the fine-tuning. Other methods require training for each target text or image from scratch, which takes at least a few minutes and in some cases hours, while our inference time per target is just a few seconds.

\section{Experiments}

\begin{figure}[t!]
    \begin{center}
    \renewcommand{\arraystretch}{0.2}
\begin{tabular}{c@{~~~}cc@{~}c@{~}c@{~}c@{~}c@{~}c@{~}c@{~}c}
\centering{  Target }
&
\centering{\begin{turn}{90}  ~~Source \end{turn}}
&
\includegraphics[width=0.095\linewidth,clip] {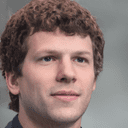}&
\includegraphics[width=0.095\linewidth,clip] {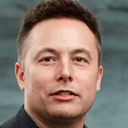}&
\includegraphics[width=0.095\linewidth,clip] {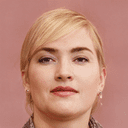}&
\includegraphics[width=0.095\linewidth,clip] {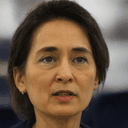}&
\includegraphics[width=0.095\linewidth,clip] {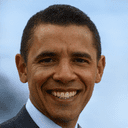}&
\includegraphics[width=0.095\linewidth,clip] {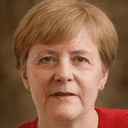}&
\includegraphics[width=0.095\linewidth,clip] {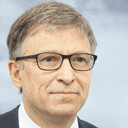}&
\\
\includegraphics[width=0.095\linewidth, clip]{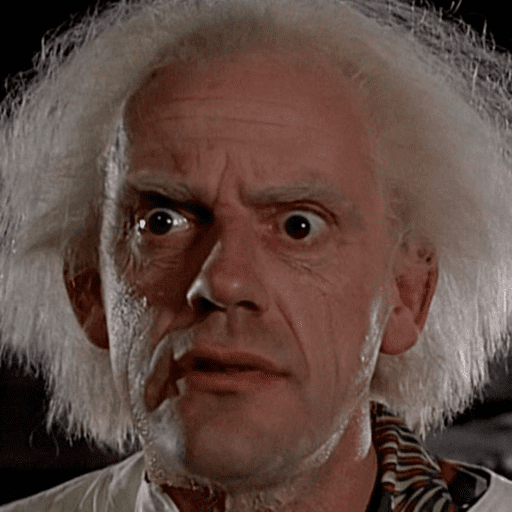}  
&
&
\includegraphics[width=0.095\linewidth,clip] {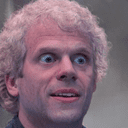}&
\includegraphics[width=0.095\linewidth,clip] {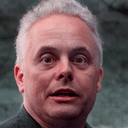}&
\includegraphics[width=0.095\linewidth,clip] {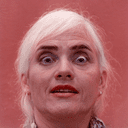}&
\includegraphics[width=0.095\linewidth,clip] {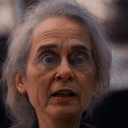}&
\includegraphics[width=0.095\linewidth,clip] {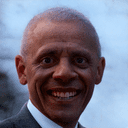}&
\includegraphics[width=0.095\linewidth,clip] {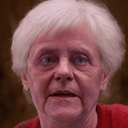}&
\includegraphics[width=0.095\linewidth,clip] {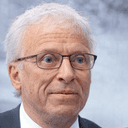}&
\\
\includegraphics[width=0.095\linewidth, clip]{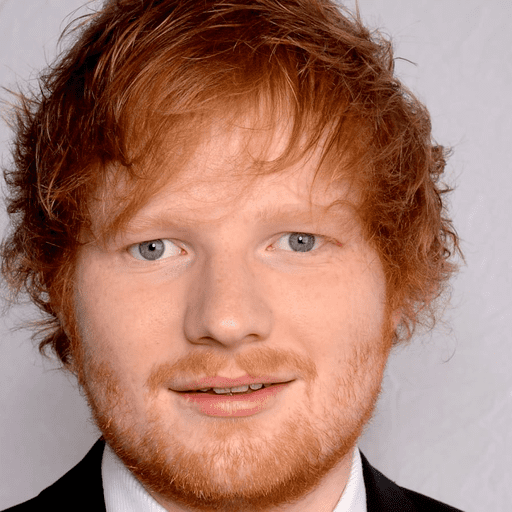}  
&
&
\includegraphics[width=0.095\linewidth,clip] {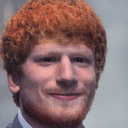} 
&
\includegraphics[width=0.095\linewidth,clip] {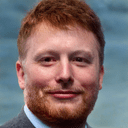}  &
\includegraphics[width=0.095\linewidth, clip]{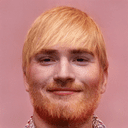}  
&
\includegraphics[width=0.095\linewidth,clip] {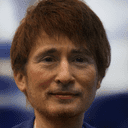} 
&
\includegraphics[width=0.095\linewidth,clip] {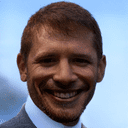}  &
\includegraphics[width=0.095\linewidth,clip] {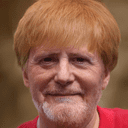}  &
\includegraphics[width=0.095\linewidth,clip] {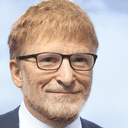}  &
\\
\includegraphics[width=0.095\linewidth, clip]{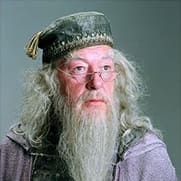} &
&
\includegraphics[width=0.095\linewidth,clip] {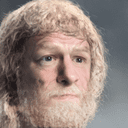} 
&
\includegraphics[width=0.095\linewidth,clip] {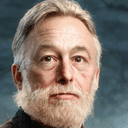}  &
\includegraphics[width=0.095\linewidth, clip]{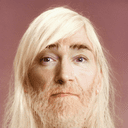}  
&
\includegraphics[width=0.095\linewidth,clip] {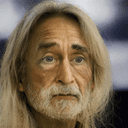} 
&
\includegraphics[width=0.095\linewidth,clip] {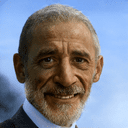}  &
\includegraphics[width=0.095\linewidth,clip] {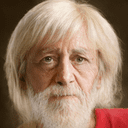}  &
\includegraphics[width=0.095\linewidth,clip] {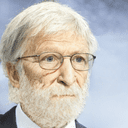}  &\\
\includegraphics[width=0.095\linewidth, clip]{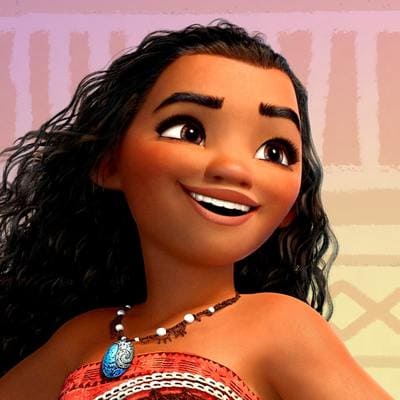}  
&
&
\includegraphics[width=0.095\linewidth,clip] {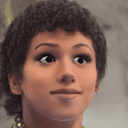} 
&
\includegraphics[width=0.095\linewidth,clip] {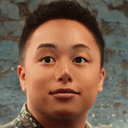}  &
\includegraphics[width=0.095\linewidth, clip]{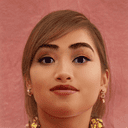}  
&
\includegraphics[width=0.095\linewidth,clip] {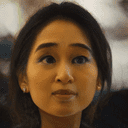} 
&
\includegraphics[width=0.095\linewidth,clip] {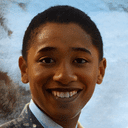}  &
\includegraphics[width=0.095\linewidth,clip] {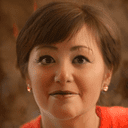}  &
\includegraphics[width=0.095\linewidth,clip] {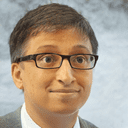}  &\\
\end{tabular}
    \captionof{figure}{
    Examples of using our optimization-based method. Output images preserve the identity of the sources, while borrowing the semantic essence of the targets. 
    }
    \label{fig:optimization}
\end{center}
    \end{figure}

We present qualitative and quantitative results that demonstrate the advantage of our method for the task of essence transfer over the most recent methods for style transfer and domain adaptation.  For a complete evaluation, we make an effort to be inclusive and compare also with methods that have somewhat different goals, i.e. text-based image editing methods.

\begin{figure}[t!]
    \begin{center}
    \renewcommand{\arraystretch}{0.2}
\begin{tabular}{c@{~~~}cc@{~}c@{~}c@{~}c@{~}c@{~}c@{~}c@{~}c}
\centering{  Target }
&
\centering{\begin{turn}{90}  ~Source \end{turn} }
&
\includegraphics[width=0.095\linewidth,clip] {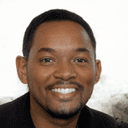}&
\includegraphics[width=0.095\linewidth,clip] {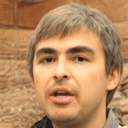}&
\includegraphics[width=0.095\linewidth,clip] {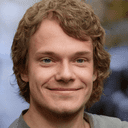}&
\includegraphics[width=0.095\linewidth,clip] {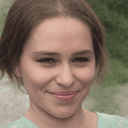}&
\includegraphics[width=0.095\linewidth,clip] {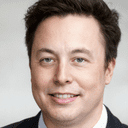}&
\includegraphics[width=0.095\linewidth,clip] {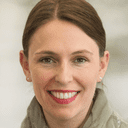}&
\includegraphics[width=0.095\linewidth,clip] {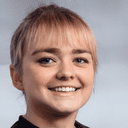}&
\\
\includegraphics[width=0.095\linewidth, clip]{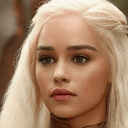}
&
&
\includegraphics[width=0.095\linewidth,clip] {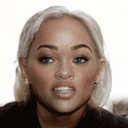}&
\includegraphics[width=0.095\linewidth,clip] {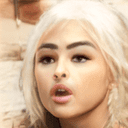}&
\includegraphics[width=0.095\linewidth,clip] {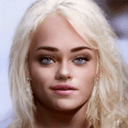}&
\includegraphics[width=0.095\linewidth,clip] {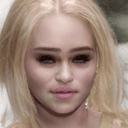}&
\includegraphics[width=0.095\linewidth,clip] {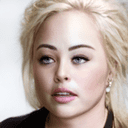}&
\includegraphics[width=0.095\linewidth,clip] {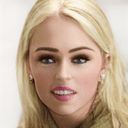}&
\includegraphics[width=0.095\linewidth,clip] {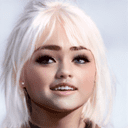}&
\\
\includegraphics[width=0.095\linewidth, clip]{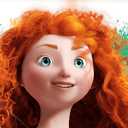}  
&
&
\includegraphics[width=0.095\linewidth,clip] {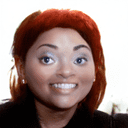}&
\includegraphics[width=0.095\linewidth,clip] {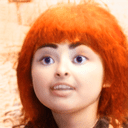}&
\includegraphics[width=0.095\linewidth,clip] {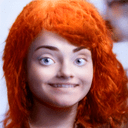}&
\includegraphics[width=0.095\linewidth,clip] {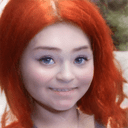}&
\includegraphics[width=0.095\linewidth,clip] {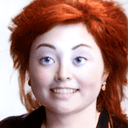}&
\includegraphics[width=0.095\linewidth,clip] {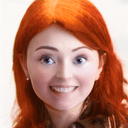}&
\includegraphics[width=0.095\linewidth,clip] {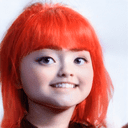}&
\\
\includegraphics[width=0.095\linewidth, clip]{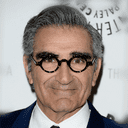}  
&
&
\includegraphics[width=0.095\linewidth,clip] {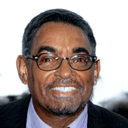}&
\includegraphics[width=0.095\linewidth,clip] {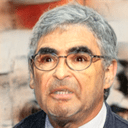}&
\includegraphics[width=0.095\linewidth,clip] {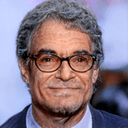}&
\includegraphics[width=0.095\linewidth,clip] {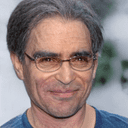}&
\includegraphics[width=0.095\linewidth,clip] {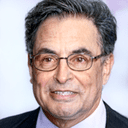}&
\includegraphics[width=0.095\linewidth,clip] {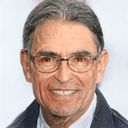}&
\includegraphics[width=0.095\linewidth,clip] {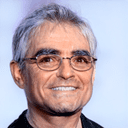}&\\
\includegraphics[width=0.095\linewidth, clip]{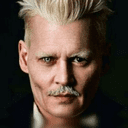}  
&
&
\includegraphics[width=0.095\linewidth,clip] {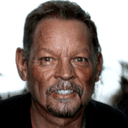}&
\includegraphics[width=0.095\linewidth,clip] {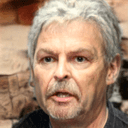}&
\includegraphics[width=0.095\linewidth,clip] {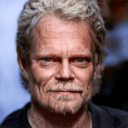}&
\includegraphics[width=0.095\linewidth,clip] {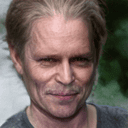}&
\includegraphics[width=0.095\linewidth,clip] {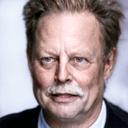}&
\includegraphics[width=0.095\linewidth,clip] {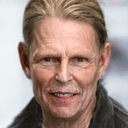}&
\includegraphics[width=0.095\linewidth,clip] {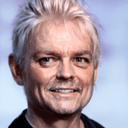}&\\
~\\
\end{tabular}
    \captionof{figure}{Examples of using our essence encoder on various targets and sources. Output images preserve the identity of the sources while borrowing the semantic essence of the targets.
    }
    \label{fig:encoder}
\end{center}
    \end{figure}

\noindent{\it Qualitative results\quad} Fig.~\ref{fig:optimization}, Fig.~\ref{fig:encoder} contain results of our optimization-based method and encoder-based method, using a wide variety of target and source images. All source images were inverted with e4e~\cite{tov2021designing}, and were not part of the training batch of sources used for optimization. The manipulation of the sources with the essence vector was done as detailed in Eq.~\ref{eq:manipulation}. We present different choices of sources and targets in Fig.~\ref{fig:optimization} and Fig.~\ref{fig:encoder}, in order to demonstrate the diversity of both our methods. For completeness, the complementary versions of the figures, in which the sources and targets in Fig.~\ref{fig:optimization} are edited with the encoder, and the sources and targets in Fig.~\ref{fig:encoder} are edited with the optimizer, are also presented in the supplementary material.

As can be seen, our essence transfer results display the most notable semantic attributes of the target. For example, when using Doc Brown as target (first row in Fig.~\ref{fig:optimization}), the signature wild, white hair is transferred from the target to all sources, as are the wide open eyes. Our methods also preserve the identity of the sources well, despite training with only $N=4$ ($N=5$) images to enforce semantic consistency for our optimization (encoder) method.  Additionally, the semantic edits are consistent across all sources, demonstrating that our method is indeed able to produce source-agnostic essence vectors. 

\begin{figure}[t]
    \begin{tabular*}{\linewidth}{@{\extracolsep{\fill}}lcc@{~~}c@{~~}c@{~~}c@{~~}c@{~~}c}
        \toprule
         &  & Quality &\multicolumn{2}{c}{Identity scores} & \multicolumn{2}{c}{Semantic scores}\\
        \cmidrule(lr){4-5}
        \cmidrule(lr){6-7}
        & &  FID ($\downarrow$) & Source ($\uparrow$)& Target ($\downarrow$) & BLIP ($\uparrow$) & CLIP ($\uparrow$)\\
        \midrule
        \multirow{7}{*}{\begin{turn}{90}{Celebrities Test} \end{turn}}
        &
        StyleGAN-NADA~\cite{gal2021stylegannada} & {\color{Orange}{215.7$\pm$26.1}} & {\color{Orange}{23.0$\pm$4.7}} & 33.0$\pm$7.1 & \textbf{84.5$\pm$3.6} & \textbf{{94.0$\pm$1.3}}  \\
        & Mind The Gap~\cite{Zhu2021MindTG} & 180.4$\pm$19.3 & {\color{Orange}{27.2$\pm$5.6}} & 39.4$\pm$8.1 &  75.8$\pm$5.6 &  75.4$\pm$7.0 \\
        & JoJoGAN~\cite{Chong2021JoJoGANOS} & 186.1$\pm$12.7  & 36.0$\pm$6.1  & {\color{Orange}{50.7$\pm$6.9}} & 72.6$\pm$7.3 & 71.8$\pm$6.2 \\
        & BlendGAN~\cite{liu2021blendgan} & 177.8$\pm$12.6  & 37.6$\pm$6.5  & \textbf{5.2$\pm$7.7} & {\color{Mahogany}{60.8$\pm$6.2}}  & {\color{Mahogany}{58.4$\pm$5.2}}  \\
        & StyleCLIP ~\cite{patashnik2021styleclip} & 166.9$\pm$9.0  & \textbf{70.7$\pm$26.0}  & 6.2$\pm$6.8 & {\color{Mahogany}54.8$\pm$6.6}   & {\color{Mahogany}55.7$\pm$5.0} \\
        & \textbf{Our encoder} & 188.7$\pm$23.2 & 39.0$\pm$6.5 & 31.9$\pm$5.7 & 69.0$\pm$6.0 & 72.6$\pm$5.5\\
        & \textbf{Our optimization} & \textbf{163.6$\pm$16.7} & 43.5$\pm$6.8 & 17.0$\pm$6.6 & 66.9$\pm$6.0  & 74.4$\pm$3.2 \\
        \midrule
        \multirow{5}{*}{\begin{turn}{90}{FFHQ Test} \end{turn}}
        &StyleGAN-NADA~\cite{gal2021stylegannada} & {\color{Orange}{220.2$\pm$41.8}} & {\color{Orange}{24.1$\pm$5.5}} & 28.3$\pm$9.2 & \textbf{81.1$\pm$4.2} & \textbf{{91.0$\pm$3.2}}  \\
        &JoJoGAN~\cite{Chong2021JoJoGANOS} & 175.2$\pm$15.2  & 42.3$\pm$4.0  & {\color{Orange}{41.7$\pm$11.4}} & 76.0$\pm$6.0 & 67.1$\pm$7.4 \\
        &BlendGAN~\cite{liu2021blendgan} & 175.1$\pm$14.5  & 37.6$\pm$5.3  & \textbf{2.4$\pm$6.0} & {\color{Mahogany}{64.4$\pm$6.7}}  & {\color{Mahogany}{54.7$\pm$7.8}}  \\
        &\textbf{Our encoder} & 175.6$\pm$23.5 & 42.5$\pm$5.5 & 30.8$\pm$6.9 & 72.8$\pm$4.9 & 66.7$\pm$6.1\\
        &\textbf{Our optimization} & \textbf{161.1$\pm$17.2} & \textbf{45.2$\pm$8.6} & 17.0$\pm$7.2 & 74.1$\pm$4.9  & 74.8$\pm$5.8 \\
        \bottomrule
    \end{tabular*}
    \captionof{table}{Quantitative comparison with baselines.
    The StyleCLIP baseline can only be applied to well-known characters (celebrities test), and Mind the Gap provides no public code at this time, thus can only be applied to the celebrities test (see main text).
    Results that indicate identity loss of the source are marked in orange; results that indicate that no semantic attributes were transferred are marked in red. 
    }
    \label{tab:celebs}
\end{figure}

\noindent{\it Quantitative Results\quad} Our experiments use as sources a set of $68$ images inverted with e4e. For each target image $I_t$ and source image $I_s$ we use our methods and the baseline methods to edit the source according to the target and produce $I_{s,t}$. We then evaluate the quality of the produced edits for each method. Since there are many works involving style transfer and domain adaptation, we focus on the most recent state of the art, including unpublished works. We focus on works that are applicable to our use-case, i.e., methods that are able to perform one-shot editing. Our baselines include BlendGAN~\cite{liu2021blendgan} and JoJoGAN~\cite{Chong2021JoJoGANOS} for face stylization, StyleGAN-NADA~\cite{gal2021stylegannada} and Mind The Gap (MTG)~\cite{Zhu2021MindTG} for domain adaptation, and as a CLIP-aided text-based image editing method, we include  StyleCLIP's~\cite{patashnik2021styleclip} global directions method. We note that since the StyleCLIP method is text-based, it can only be used in manipulations where the target is a well-known character. Despite its inherent limitation, we also present this comparison for completeness, since the global directions method resembles ours in that it outputs a target-dependent and source-agnostic direction in the StyleGAN $\mathcal{S}$ space to perform a manipulation according to an input textual description. Since our methods strive to transfer the features of the high-level textual description of the target, we find this comparison to be relevant as well.   

The goal of essence transfer is twofold. First, we wish to transfer the semantic properties that constitute the high-level textual description from a target image $I_t$ to a source image $I_s$. Second, we wish to maintain the identity of $I_s$ as much as possible. We therefore suggest two types of metrics to evaluate the quality of a proposed essence transfer result, $I_{s,t}$. The first type employs the ArcFace~\cite{Deng2019ArcFaceAA} network for face recognition to ensure that the manipulation maintains the identity of $I_s$ as much as possible, while avoiding an identity shift towards $I_t$, i.e. we calculate:
\begin{align*}
    {\text{ID-score}_{source}}(I_{s,t})= \langle R(I_s), R(I_{s,t}) \rangle, \text{   } {\text{ID-score}_{target}}(I_{s,t})= \langle R(I_t), R(I_{s,t}) \rangle,
\end{align*}
where $R$ denotes a pre-trained ArcFace face recognition representation, and $\langle\cdot,\cdot\rangle$ computes cosine similarity. Since neither of our methods uses face recognition in the training process, this metric faithfully measures how well our manipulations preserve the source identity. Intuitively, since we add semantic features from the target, we shift the identity of the source to some extent. For example, modifying the gender of the source induces an inherent change in one of the identity attributes of the source. The combination of scores ${\text{ID-score}_{source}}, {\text{ID-score}_{target}}$ reveals whether the manipulation was able to remain close to the identity of the source or shifted toward the identity of the target. A successful essence transfer is expected to maintain a \emph{high} ${\text{ID-score}_{source}}$  score, and a \emph{low} ${\text{ID-score}_{target}}$ score, indicating that the manipulation's identity fits the source better than the target. 

Next, to estimate the semantic quality of the manipulation, we use the latent spaces of BLIP~\cite{li2022blip} and CLIP~\cite{radford2021learning}, as follows:
\begin{align*}
    \mathcal{\text{Semantic-score}}(I_{s,t})=\langle C(I_t), C(I_{s,t}) \rangle,
\end{align*}
where $C$ notates a pre-trained BLIP or CLIP image encoder, and $\langle\cdot,\cdot\rangle$ computes cosine similarity.
Since our method, as well as most baselines~\cite{gal2021stylegannada,Zhu2021MindTG,patashnik2021styleclip}, use the latent space of CLIP in the training process, BLIP provides an important alternative for estimating the semantic similarity between the target image $I_t$ and the manipulation $I_{s,t}$.
For each target $I_t$, the overall identity scores and semantic scores are calculated as an average of the scores for all source images. The aggregated scores for the models are calculated as an average of the score for each target, i.e. we average the results across the sources for each target, and then average across the targets to obtain the model's final score. We also present the standard deviations as an indication of the method's consistency. 

Additionally, in accordance with previous works on style transfer, we present the Fr\'echet inception distance (FID)~\cite{fid} as implemented in~\cite{Seitzer2020FID} to estimate the quality of the manipulations, which is calculated as follows: 
\begin{align*}
    \mathcal{\text{FID-score}}(I_{s_1,t}, ..., I_{s_{68},t})=\text{FID}(\{I_{s_1,t}, ..., I_{s_{68},t}\}, \{I_1,..., I_{7,000}\}),
\end{align*}
where $\{I_{s_1,t}, ..., I_{s_{68},t}\}$ are the manipulations of the sources induced by the target $t$, and $\{I_1,..., I_{7,000}\}$ is a set of $7,000$ randomly chosen images from the FFHQ~\cite{Karras2019ASG} dataset, which provides the background distribution of natural faces.
Since some of our baselines are trained to adapt the domain of the target, we calculate the FID score only for the targets describing a human face, in order to avoid biasing the results against these baselines. Note that this calculation produces a relatively high FID score for all methods, since the produced dataset $\{I_{s_1,t}, ..., I_{s_{68},t}\}$ is inherently limited in its diversity, due to the fact that all images share semantic properties transferred from $t$, leading to a shift from the distribution of unedited faces, which are more diverse. However, methods that suffer from mode collapse or overfitting are expected to achieve a much higher (lower is better) score than those that preserve the original identity of the source images, since identity preservation will lead to greater diversity among the results.

\begin{figure}[t!]
    \begin{center}
    \renewcommand{\arraystretch}{0.2}
\begin{tabular}{c@{~~}ccccc@{~~}cccc@{~~}cccc}
\centering{  Target }
&
\centering{ \begin{turn}{90}~ Src. \end{turn} }
&
\includegraphics[width=0.065\linewidth,clip] {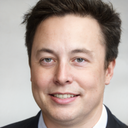}&
\includegraphics[width=0.065\linewidth,clip] {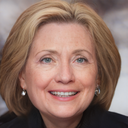}&
\includegraphics[width=0.065\linewidth,clip] {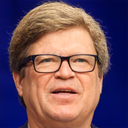}&
\includegraphics[width=0.065\linewidth,clip] {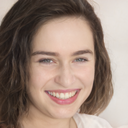}&
\includegraphics[width=0.065\linewidth,clip] {underfit2/4.png}&
\includegraphics[width=0.065\linewidth,clip] {underfit2/16.png}&
\includegraphics[width=0.065\linewidth,clip] {underfit2/18.png}&
\includegraphics[width=0.065\linewidth,clip] {underfit2/42.png}&
\includegraphics[width=0.065\linewidth,clip] {underfit2/4.png}&
\includegraphics[width=0.065\linewidth,clip] {underfit2/16.png}&
\includegraphics[width=0.065\linewidth,clip] {underfit2/18.png}&
\includegraphics[width=0.065\linewidth,clip] {underfit2/42.png}
\\
\includegraphics[width=0.065\linewidth, clip]{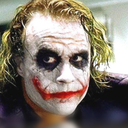}
&
&
\includegraphics[width=0.065\linewidth,clip] {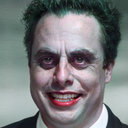} 
&
\includegraphics[width=0.065\linewidth,clip] {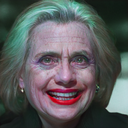} &
\includegraphics[width=0.065\linewidth, clip]{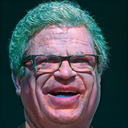}  
&
\includegraphics[width=0.065\linewidth,clip] {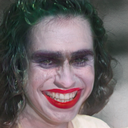} 
&
\includegraphics[width=0.065\linewidth,clip] {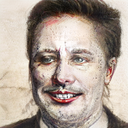} 
&
\includegraphics[width=0.065\linewidth,clip] {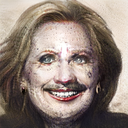} &
\includegraphics[width=0.065\linewidth, clip]{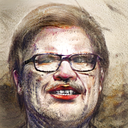}  
&
\includegraphics[width=0.065\linewidth,clip] {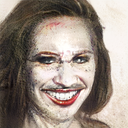} 
&
\includegraphics[width=0.065\linewidth,clip] {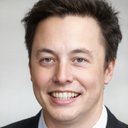} 
&
\includegraphics[width=0.065\linewidth,clip] {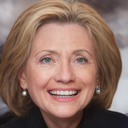} &
\includegraphics[width=0.065\linewidth, clip]{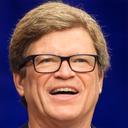}  
&
\includegraphics[width=0.065\linewidth,clip] {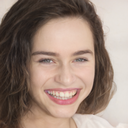}
\\
\includegraphics[width=0.065\linewidth, clip]{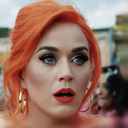}
&
&
\includegraphics[width=0.065\linewidth,clip] {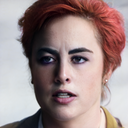} 
&
\includegraphics[width=0.065\linewidth,clip] {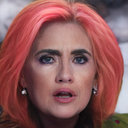} &
\includegraphics[width=0.065\linewidth, clip]{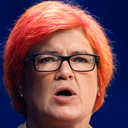}  
&
\includegraphics[width=0.065\linewidth,clip] {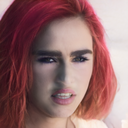} 
&
\includegraphics[width=0.065\linewidth,clip] {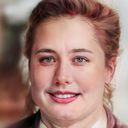} 
&
\includegraphics[width=0.065\linewidth,clip] {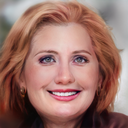} &
\includegraphics[width=0.065\linewidth, clip]{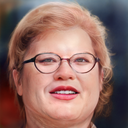}  
&
\includegraphics[width=0.065\linewidth,clip] {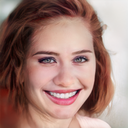} 
&
\includegraphics[width=0.065\linewidth,clip] {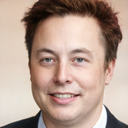} 
&
\includegraphics[width=0.065\linewidth,clip] {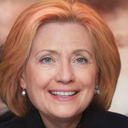} &
\includegraphics[width=0.065\linewidth, clip]{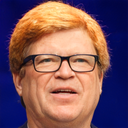}  
&
\includegraphics[width=0.065\linewidth,clip] {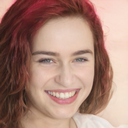}
\\
\includegraphics[width=0.065\linewidth, clip]{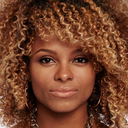}
&
&
\includegraphics[width=0.065\linewidth,clip] {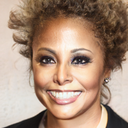} 
&
\includegraphics[width=0.065\linewidth,clip] {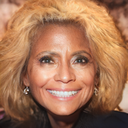} &
\includegraphics[width=0.065\linewidth, clip]{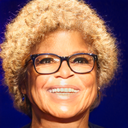}  
&
\includegraphics[width=0.065\linewidth,clip] {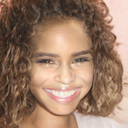} 
&
\includegraphics[width=0.065\linewidth,clip] {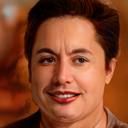} 
&
\includegraphics[width=0.065\linewidth,clip] {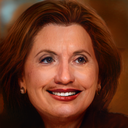} &
\includegraphics[width=0.065\linewidth, clip]{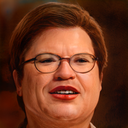}  
&
\includegraphics[width=0.065\linewidth,clip] {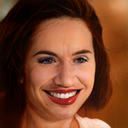} 
&
\includegraphics[width=0.065\linewidth,clip] {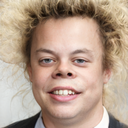} 
&
\includegraphics[width=0.065\linewidth,clip] {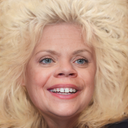} &
\includegraphics[width=0.065\linewidth, clip]{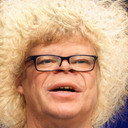}  
&
\includegraphics[width=0.065\linewidth,clip] {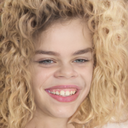}
\\
\includegraphics[width=0.065\linewidth, clip]{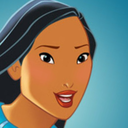}
&
&
\includegraphics[width=0.065\linewidth,clip] {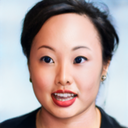} 
&
\includegraphics[width=0.065\linewidth,clip] {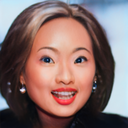} &
\includegraphics[width=0.065\linewidth, clip]{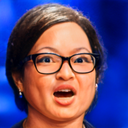}  
&
\includegraphics[width=0.065\linewidth,clip] {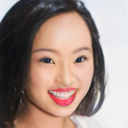} 
&
\includegraphics[width=0.065\linewidth,clip] {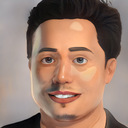}
&
\includegraphics[width=0.065\linewidth,clip] {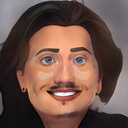} &
\includegraphics[width=0.065\linewidth, clip]{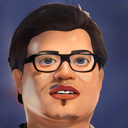}  
&
\includegraphics[width=0.065\linewidth,clip] {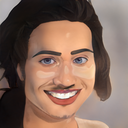} 
&
\includegraphics[width=0.065\linewidth,clip] {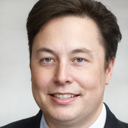} 
&
\includegraphics[width=0.065\linewidth,clip] {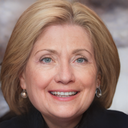} &
\includegraphics[width=0.065\linewidth, clip]{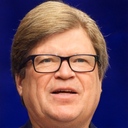} 
&
\includegraphics[width=0.065\linewidth,clip] {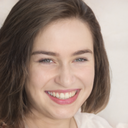}
\\
\includegraphics[width=0.065\linewidth, clip]{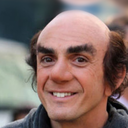}
&
&
\includegraphics[width=0.065\linewidth,clip] {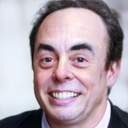} 
&
\includegraphics[width=0.065\linewidth,clip] {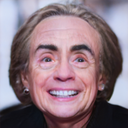} &
\includegraphics[width=0.065\linewidth, clip]{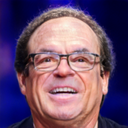}  
&
\includegraphics[width=0.065\linewidth,clip] {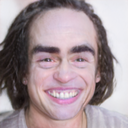} 
&
\includegraphics[width=0.065\linewidth,clip] {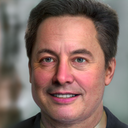} 
&
\includegraphics[width=0.065\linewidth,clip] {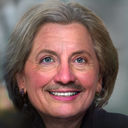} &
\includegraphics[width=0.065\linewidth, clip]{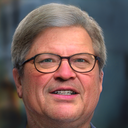}  
&
\includegraphics[width=0.065\linewidth,clip] {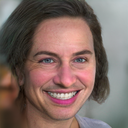} 
&
\includegraphics[width=0.065\linewidth,clip] {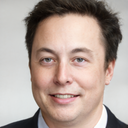} 
&
\includegraphics[width=0.065\linewidth,clip] {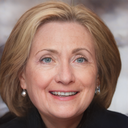} &
\includegraphics[width=0.065\linewidth, clip]{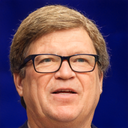}  
&
\includegraphics[width=0.065\linewidth,clip] {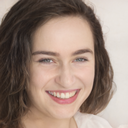}
\\
\includegraphics[width=0.065\linewidth, clip]{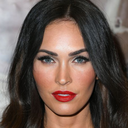}
&
&
\includegraphics[width=0.065\linewidth,clip] {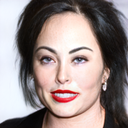} 
&
\includegraphics[width=0.065\linewidth,clip] {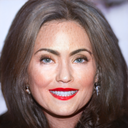} &
\includegraphics[width=0.065\linewidth, clip]{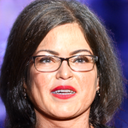}  
&
\includegraphics[width=0.065\linewidth,clip] {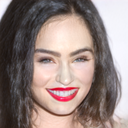} 
&
\includegraphics[width=0.065\linewidth,clip] {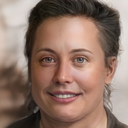} 
&
\includegraphics[width=0.065\linewidth,clip] {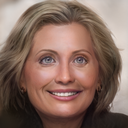} &
\includegraphics[width=0.065\linewidth, clip]{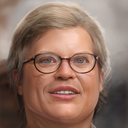}  
&
\includegraphics[width=0.065\linewidth,clip] {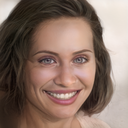} 
&
\includegraphics[width=0.065\linewidth,clip] {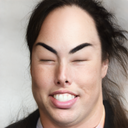} 
&
\includegraphics[width=0.065\linewidth,clip] {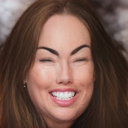} &
\includegraphics[width=0.065\linewidth, clip]{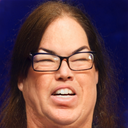}  
&
\includegraphics[width=0.065\linewidth,clip] {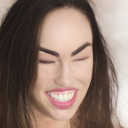}
\\\\
 & &  & \multicolumn{2}{c}{Ours} & & &   \multicolumn{2}{c}{BlendGAN}& & \multicolumn{4}{c}{StyleCLIP}\\
\end{tabular}
    \captionof{figure}{Comparison to methods that only partially transfer the semantic properties. First three rows are manipulations with our optimizer,
    and the last three are with our encoder. 
    }
    \label{fig:underfit}
\end{center}
    \end{figure}

\begin{figure}[t!]
    \begin{center}
    \renewcommand{\arraystretch}{0.2}
\begin{tabular}{c@{~~}ccccc@{~~}cccc@{~~}cccc}
\centering{  Target }
&
\centering{ \begin{turn}{90}~ Src. \end{turn} }
&
\includegraphics[width=0.065\linewidth,clip] {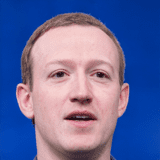}&
\includegraphics[width=0.065\linewidth,clip] {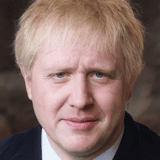}&
\includegraphics[width=0.065\linewidth,clip] {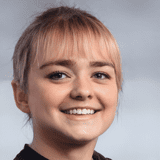}&
\includegraphics[width=0.065\linewidth,clip] {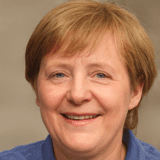}&
\includegraphics[width=0.065\linewidth,clip] {overfit/3.png}&
\includegraphics[width=0.065\linewidth,clip] {overfit/15.png}&
\includegraphics[width=0.065\linewidth,clip] {overfit/44.png}&
\includegraphics[width=0.065\linewidth,clip] {overfit/62.png}&
\includegraphics[width=0.065\linewidth,clip] {overfit/3.png}&
\includegraphics[width=0.065\linewidth,clip] {overfit/15.png}&
\includegraphics[width=0.065\linewidth,clip] {overfit/44.png}&
\includegraphics[width=0.065\linewidth,clip] {overfit/62.png}
\\
\includegraphics[width=0.065\linewidth, clip]{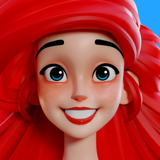}
&
&
\includegraphics[width=0.065\linewidth,clip] {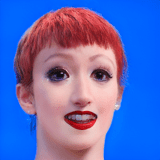} 
&
\includegraphics[width=0.065\linewidth,clip] {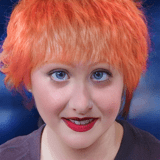} &
\includegraphics[width=0.065\linewidth, clip]{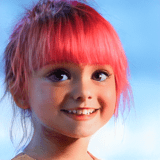}  
&
\includegraphics[width=0.065\linewidth,clip] {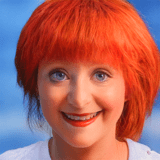} 
&
\includegraphics[width=0.065\linewidth,clip] {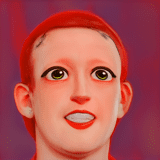} 
&
\includegraphics[width=0.065\linewidth,clip] {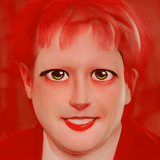} &
\includegraphics[width=0.065\linewidth, clip]{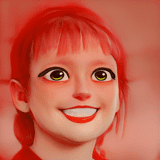}  
&
\includegraphics[width=0.065\linewidth,clip] {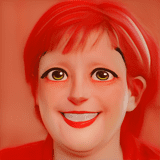} 
&
\includegraphics[width=0.065\linewidth,clip] {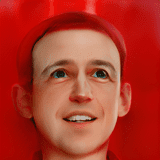} 
&
\includegraphics[width=0.065\linewidth,clip] {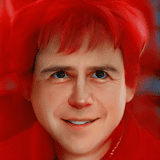} &
\includegraphics[width=0.065\linewidth, clip]{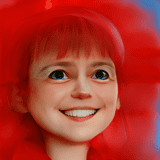}  
&
\includegraphics[width=0.065\linewidth,clip] {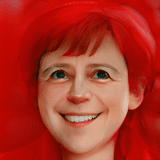}
\\
\includegraphics[width=0.065\linewidth, clip]{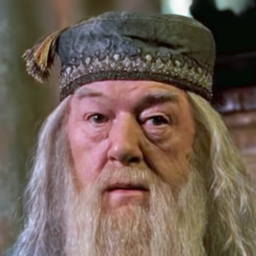}
&
&
\includegraphics[width=0.065\linewidth,clip] {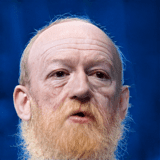} 
&
\includegraphics[width=0.065\linewidth,clip] {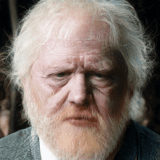} &
\includegraphics[width=0.065\linewidth, clip]{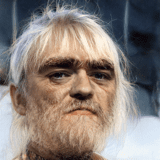}  
&
\includegraphics[width=0.065\linewidth,clip] {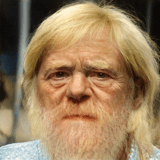} 
&
\includegraphics[width=0.065\linewidth,clip] {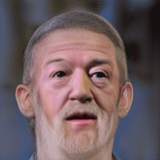} 
&
\includegraphics[width=0.065\linewidth,clip] {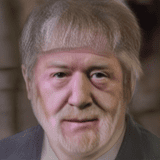} &
\includegraphics[width=0.065\linewidth, clip]{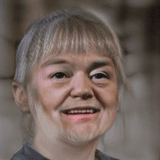}  
&
\includegraphics[width=0.065\linewidth,clip] {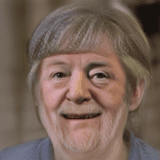} 
&
\includegraphics[width=0.065\linewidth,clip] {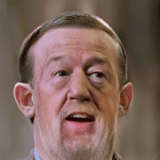} 
&
\includegraphics[width=0.065\linewidth,clip] {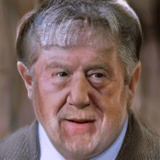} &
\includegraphics[width=0.065\linewidth, clip]{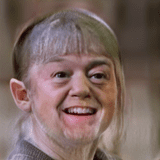}  
&
\includegraphics[width=0.065\linewidth,clip] {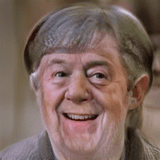}
\\
\includegraphics[width=0.065\linewidth, clip]{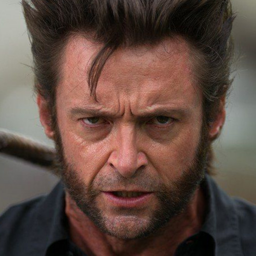}
&
&
\includegraphics[width=0.065\linewidth,clip] {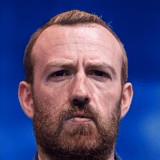} 
&
\includegraphics[width=0.065\linewidth,clip] {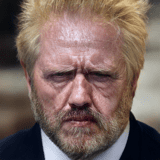} &
\includegraphics[width=0.065\linewidth, clip]{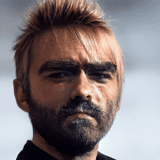}  
&
\includegraphics[width=0.065\linewidth,clip] {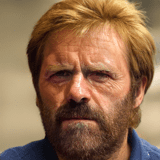} 
&
\includegraphics[width=0.065\linewidth,clip] {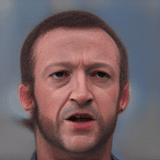} 
&
\includegraphics[width=0.065\linewidth,clip] {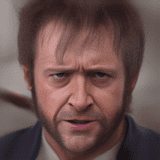} &
\includegraphics[width=0.065\linewidth, clip]{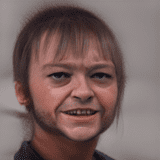}  
&
\includegraphics[width=0.065\linewidth,clip] {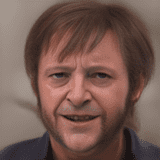} 
&
\includegraphics[width=0.065\linewidth,clip] {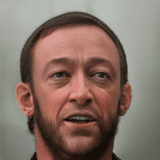} 
&
\includegraphics[width=0.065\linewidth,clip] {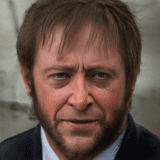} &
\includegraphics[width=0.065\linewidth, clip]{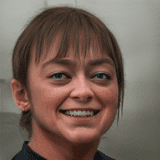}  
&
\includegraphics[width=0.065\linewidth,clip] {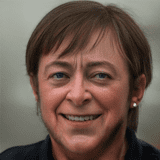}
\\
\includegraphics[width=0.065\linewidth, clip]{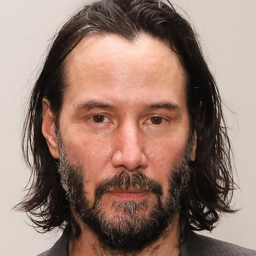}
&
&
\includegraphics[width=0.065\linewidth,clip] {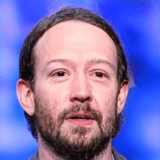} 
&
\includegraphics[width=0.065\linewidth,clip] {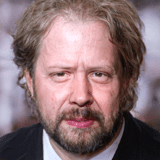} &
\includegraphics[width=0.065\linewidth, clip]{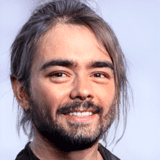}  
&
\includegraphics[width=0.065\linewidth,clip] {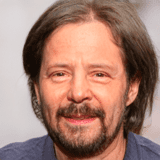} 
&
\includegraphics[width=0.065\linewidth,clip] {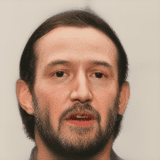} 
&
\includegraphics[width=0.065\linewidth,clip] {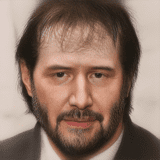} &
\includegraphics[width=0.065\linewidth, clip]{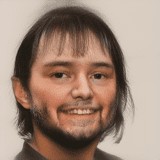}  
&
\includegraphics[width=0.065\linewidth,clip] {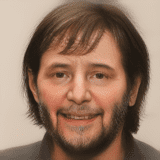} 
&
\includegraphics[width=0.065\linewidth,clip] {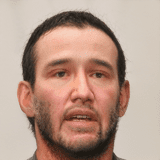} 
&
\includegraphics[width=0.065\linewidth,clip] {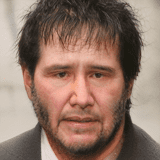} &
\includegraphics[width=0.065\linewidth, clip]{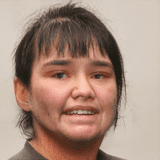}  
&
\includegraphics[width=0.065\linewidth,clip] {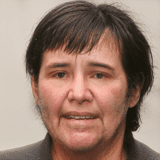}
\\
\includegraphics[width=0.065\linewidth, clip]{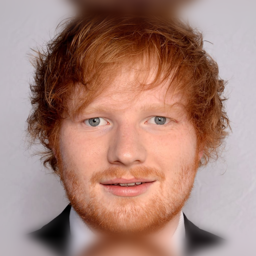}
&
&
\includegraphics[width=0.065\linewidth,clip] {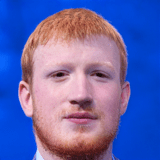} 
&
\includegraphics[width=0.065\linewidth,clip] {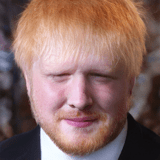} &
\includegraphics[width=0.065\linewidth, clip]{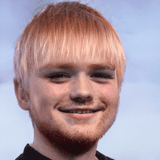}  
&
\includegraphics[width=0.065\linewidth,clip] {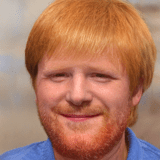} 
&
\includegraphics[width=0.065\linewidth,clip] {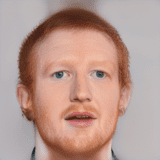} 
&
\includegraphics[width=0.065\linewidth,clip] {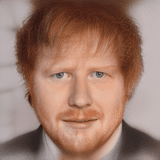} &
\includegraphics[width=0.065\linewidth, clip]{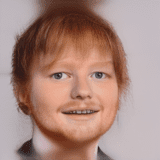}  
&
\includegraphics[width=0.065\linewidth,clip] {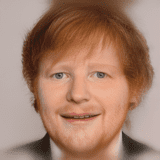} 
&
\includegraphics[width=0.065\linewidth,clip] {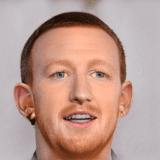} 
&
\includegraphics[width=0.065\linewidth,clip] {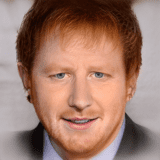} &
\includegraphics[width=0.065\linewidth, clip]{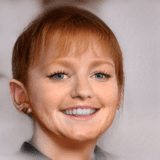}  
&
\includegraphics[width=0.065\linewidth,clip] {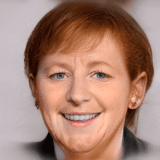}
\\
\includegraphics[width=0.065\linewidth, clip]{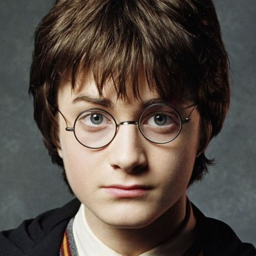}
&
&
\includegraphics[width=0.065\linewidth,clip] {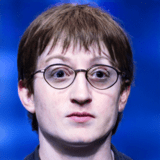} 
&
\includegraphics[width=0.065\linewidth,clip] {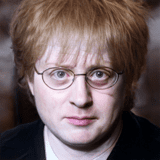} &
\includegraphics[width=0.065\linewidth, clip]{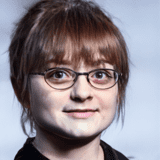}  
&
\includegraphics[width=0.065\linewidth,clip] {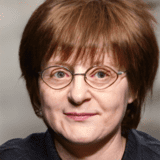} 
&
\includegraphics[width=0.065\linewidth,clip] {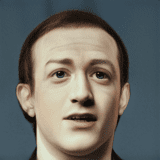} 
&
\includegraphics[width=0.065\linewidth,clip] {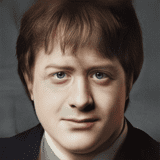} &
\includegraphics[width=0.065\linewidth, clip]{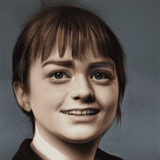}  
&
\includegraphics[width=0.065\linewidth,clip] {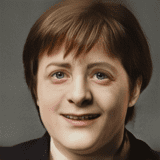} 
&
\includegraphics[width=0.065\linewidth,clip] {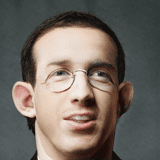} 
&
\includegraphics[width=0.065\linewidth,clip] {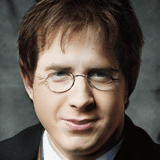} &
\includegraphics[width=0.065\linewidth, clip]{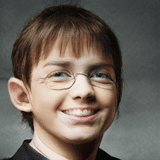}  
&
\includegraphics[width=0.065\linewidth,clip] {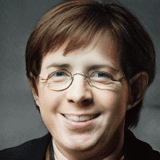}
\\\\
 & & & \multicolumn{2}{c}{Ours} & & &   \multicolumn{2}{c}{JoJoGAN}& &  \multicolumn{4}{c}{Mind The Gap}\\
\end{tabular}
    \captionof{figure}{Comparison to methods that suffer from high loss of source identity. First three rows are manipulations with our optimizer, and last three with our encoder.
    }
    \label{fig:overfit}
\end{center}
    \end{figure}

We present two experiments. For the first, we construct a comparison in a setting that is more similar to the setting the baselines were trained for, i.e. we construct a dataset of $31$ images of celebrity faces with notable or extreme semantic properties, such as unusual hair colors and styles, beards, glasses, as well as a variety of ages, genders, and ethnicities, and also out-of-domain animated characters (see the supplementary material for all examples used in this experiment). For the text-based baseline, we employ the same course of action as in the StyleCLIP paper, where the textual prompt for the manipulation is of the form ``an image of \{\emph{name of target}\}''. Our second experiment involves targets with less extreme semantic features. We use the first $50$ images of the FFHQ~\cite{Karras2019ASG} dataset as targets, and the same $68$ source images as before. Since our targets are no longer well-known characters, the baseline for text-based image editing is no longer applicable. 
Additionally, for the Mind The Gap baseline~\cite{Zhu2021MindTG}, no official code was released- although the authors kindly provided results for the first experiment, but not the second one -  therefore this baseline is not presented in our second experiment.

Tab.~\ref{tab:celebs} presents the results of both our experiments.
We divide the methods into three types. (i) Methods that demonstrate underfitting, i.e., fail to transfer the essence of the target. These methods perform very well on the identity metrics and very poorly on the semantic metrics. As can be seen in Tab.~\ref{tab:celebs}, both BlendGAN and the StyleCLIP demonstrate this phenomenon. Marked in red in the tables are the similarity scores for the methods by BLIP and CLIP. Both are significantly lower than the semantic scores of the other methods. See Fig.~\ref{fig:underfit} for examples of this case from our first experiment. BlendGAN focuses on modifying almost only the colors, and StyleCLIP either hardly changes the semantic properties or distorts the sources. (ii) methods that demonstrate overfitting, i.e. methods that suffer from identity preservation issues. These methods transfer most or all of the semantic features of the target, and eliminate the source identity in the process or create a blended identity of source and target. This results in very high semantic compatibility scores, but on the other hand, a failure in identity preservation. As can be expected, the methods designed for domain adaptation, i.e., StyleGAN-NADA and Mind The Gap fall in this category, as does JoJoGAN. The values marked in orange in Tab.~\ref{tab:celebs} demonstrate that both StyleGAN-NADA and Mind The Gap obtain very low source identity scores (significantly lower than the other methods), while JoJoGAN receives the highest (lower is better) target identity score in both experiments ($50.7\%$ on the celebrities test and $41.7\%$ on the FFHQ test, surpassing all other baselines by more than $10\%$). This indicates that StyleGAN-NADA and Mind The Gap fall short on identity preservation, while JoJoGAN results in an image derived from the identity of the target instead of the source. For example, the Ariel target (first row in Fig.~\ref{fig:overfit}) demonstrates that the baselines result in a unified identity with the semantic features of the target. Similarly, the Keanu Reeves and Ed Sheeran targets (fourth and fifth rows in Fig.~\ref{fig:overfit}) result in blended identities with the baselines. We omit StyleGAN-NADA from Fig.~\ref{fig:overfit} for brevity, as the other two methods scored higher in terms of identity preservation. The full comparison, as well as comparisons from our second experiment can be found in the supplementary material.

Lastly, (iii) methods that successfully transfer the semantic properties of the target (have high BLIP and CLIP similarity scores) while also preserving the identity of the source more than the target (i.e., $\text{ID-score}_{source} > \text{ID-score}_{target}$), which both our methods fall under. When analyzing the quality of the manipulated images, our optimization-based method scores the best overall FID score in both experiments by a significant margin, indicating that it is able to produce high-quality manipulations. In addition, our optimization demonstrates a very low (lower is better) target identity score, suggesting that our essence transfer does not borrow from the identity of the target in order to obtain the semantic changes. While our encoder preserves the identity quality ($\text{ID-score}_{source} > \text{ID-score}_{target}$), notice that it achieves a higher target identity score, indicating that our encoder is not as successful as our optimizer in identity preservation. This can be attributed to two facts. First, our encoder is based on a pre-trained inversion encoder that encapsulates the target identity by design. Second, while our optimizer learns an essence vector for each target, the encoder is only fine-tuned on a small set of images and is not optimized for each target at inference time. 

\begin{figure}[t!]
    \begin{center}
    \renewcommand{\arraystretch}{0.2}
\includegraphics[width=0.85\linewidth,clip] {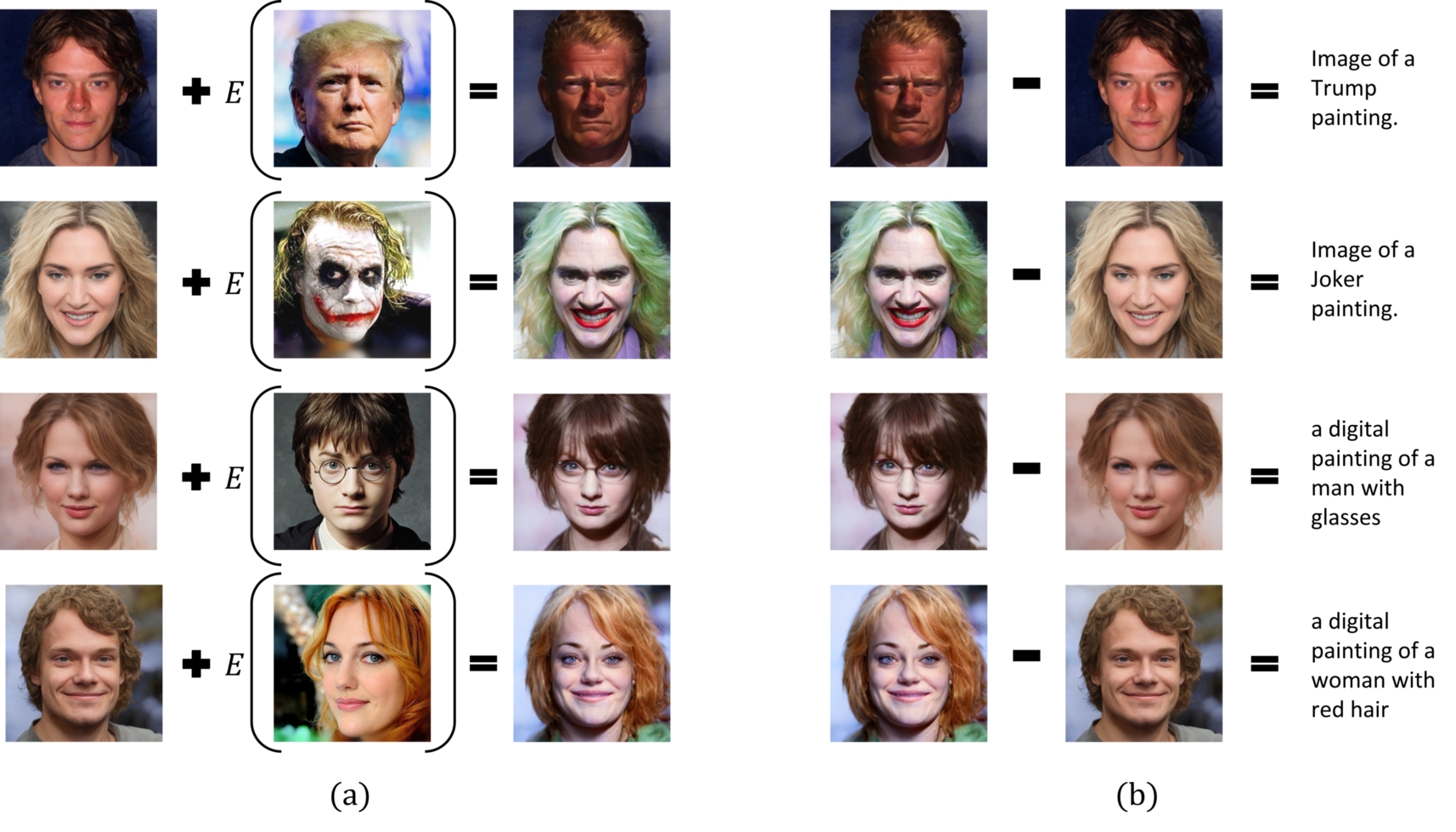} 
    \captionof{figure}{Examples of essence decoding. (a) presents the targets, sources, and manipulation results, with $E$ representing the essence extraction, i.e. we add the essence of the right image to the left image, and (b) demonstrates the decoding of the essence vectors for each example. 
    }
    \label{fig:interpret}
\end{center}
    \end{figure}
\noindent{\it Essence Interpretability\quad}
To demonstrate that our methods indeed produce essence vectors that correspond to the semantic attributes of the target image $I_t$, we present results of decoding the essence vectors to text.
We observe semantic differences by applying a decoder on the vector $d = C(I_{s,t})-C(I_s)$ where $C$ represents the semantic image encoder of CLIP or BLIP, i.e. we decode the difference between the source image after and before the manipulation.  Fig.~\ref{fig:interpret} shows examples for four different edits and their interpretations. The Donald Trump and the Joker edits (first and second row of Fig.~\ref{fig:interpret}) were performed with our optimization-based method, encoded with CLIP, and decoded with~\cite{Tewel2021ZeroShotIG}, and the rest were performed using our encoder approach and encoded/decoded with BLIP. As can be seen, the textual interpretations of each direction correspond well to the semantic properties of the targets, and for the Donald Trump and Joker edits, the directions are decoded as Trump and the Joker, demonstrating the ability of our method to capture essential semantic features of the targets used. For targets with less distinct semantic properties, the decoding shows that the apparent gender is transferred along with other significant semantic properties such as hair color and eye glasses. See the supplementary material for more examples of essence interpretability using decoding.

\begin{figure}[t!]
    \begin{center}
    \renewcommand{\arraystretch}{0.2}
\begin{tabular}{c@{~~}cc@{~}c@{~}c@{~}c@{~~~~~}c@{~}c@{~}c@{~}c@{~}c}
\centering{  Target }
&
\centering{ \begin{turn}{90}~~ Src. \end{turn} }
&
\includegraphics[width=0.08\linewidth,clip] {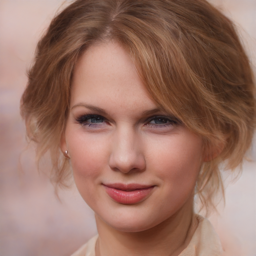}&
\includegraphics[width=0.08\linewidth,clip] {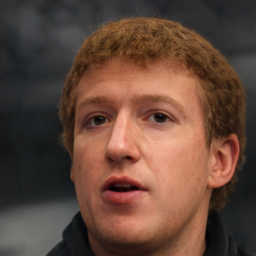}&
\includegraphics[width=0.08\linewidth,clip] {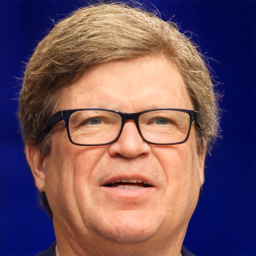}&
\includegraphics[width=0.08\linewidth,clip] {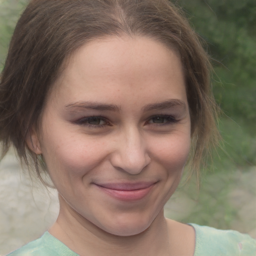}&
\includegraphics[width=0.08\linewidth,clip] {limitations2/1.png}&
\includegraphics[width=0.08\linewidth,clip] {limitations2/2.png}&
\includegraphics[width=0.08\linewidth,clip] {limitations2/3.png}&
\includegraphics[width=0.08\linewidth,clip] {limitations2/4.png}&
\\
\includegraphics[width=0.08\linewidth, clip]{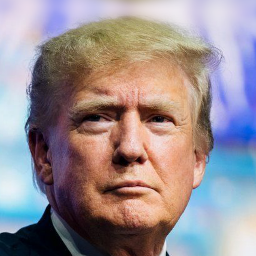}
&
&
\includegraphics[width=0.08\linewidth,clip] {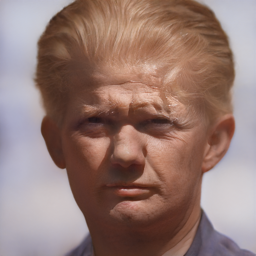} 
&
\includegraphics[width=0.08\linewidth,clip] {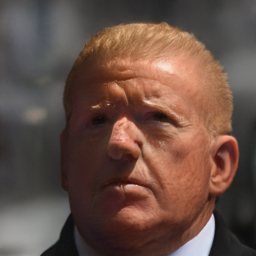} &
\includegraphics[width=0.08\linewidth, clip]{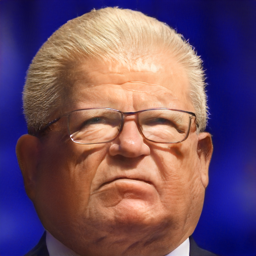}  
&
\includegraphics[width=0.08\linewidth,clip] {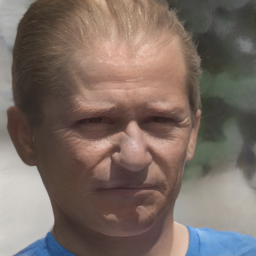} 
&
\includegraphics[width=0.08\linewidth,clip] {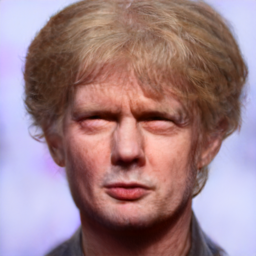} 
&
\includegraphics[width=0.08\linewidth,clip] {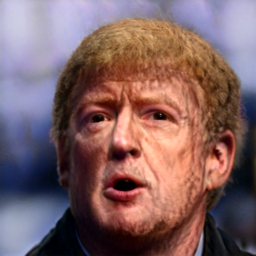} &
\includegraphics[width=0.08\linewidth, clip]{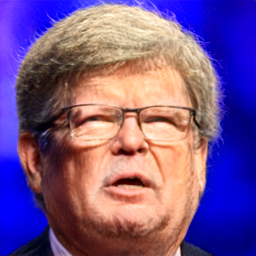}  
&
\includegraphics[width=0.08\linewidth,clip] {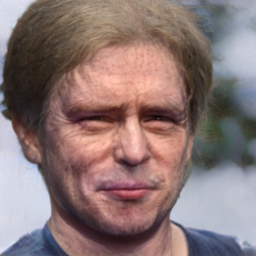} 
\\
\includegraphics[width=0.08\linewidth, clip]{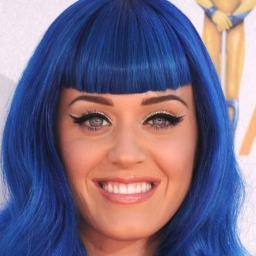}  
&
&
\includegraphics[width=0.08\linewidth,clip] {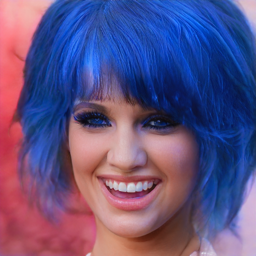} 
&
\includegraphics[width=0.08\linewidth,clip] {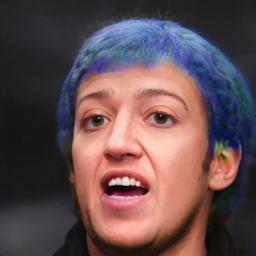} &
\includegraphics[width=0.08\linewidth, clip]{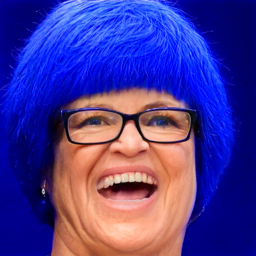}  
&
\includegraphics[width=0.08\linewidth,clip] {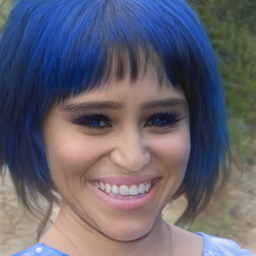} 
&
\includegraphics[width=0.08\linewidth,clip] {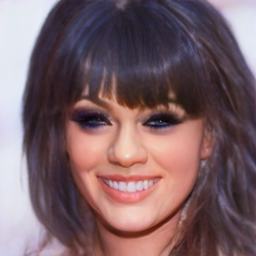} 
&
\includegraphics[width=0.08\linewidth,clip] {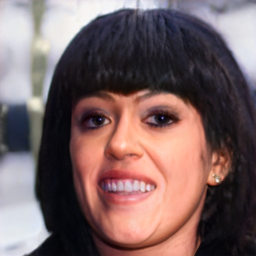} &
\includegraphics[width=0.08\linewidth, clip]{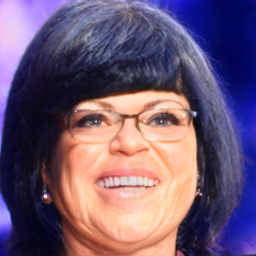}  
&
\includegraphics[width=0.08\linewidth,clip] {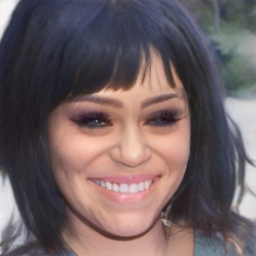} \\
~\\
 & & &\multicolumn{2}{c}{(a)} & &  & \multicolumn{2}{c}{(b)}\\
\end{tabular}
    \captionof{figure}{A comparison between our optimizer (a), and encoder (b). 
    }
    \label{fig:limitations}
\end{center}
    \end{figure}

\noindent{\it Limitations of the encoder-based approach\quad} Unlike the optimizer, the encoder is not re-trained for each target. This results in an accuracy-runtime trade-off, i.e., while the encoder produces an essence vector in a few seconds, in some cases - where the target contains unconventional semantic properties - it produces a result that does not encapsulate all the semantic attributes one would expect to be included in the essence. In contrast, since the optimization is performed from scratch for each target, it takes longer (a few minutes) to produce the result, but it is more accurate.  Fig.~\ref{fig:limitations} presents two examples of such challenging targets, where the optimization-based method is superior to the encoder. For Donald Trump (first row in Fig.~\ref{fig:limitations}), optimization results in an essence that includes all notable semantic properties- the wrinkles, lips, and unique hair color - while the result of the encoder fails to capture the unique attributes with the same accuracy. Similarly, for Katy Perry (second row in Fig.~\ref{fig:limitations}), optimization captures the unconventional hair color, while the encoder fails to do so. 
As evident from Tab.~\ref{tab:celebs}, while both the encoder and optimizer receive high semantic scores, the optimizer allows for results with higher quality (lower FID, lower target ID score).

\noindent{\it Ablation Study\quad} We refer the reader to the supplementary material for an ablation study that examines the impact of each loss term of our method.

\section{Conclusions}
We define a novel task referred to as essence transfer. Unlike style transfer or domain adaptation, essence transfer draws semantic features that correspond to the high-level textual description of an image. Essence transfer is particularly challenging since the set of attributes that constitute the high-level description may differ from image to image. We propose an optimizer and an encoder, both based on double-additivity in the latent spaces of StyleGAN and CLIP, and measure our method against state of the art methods adapted from style transfer and domain adaptation. Our extensive experiments demonstrate that our novel formulation is significantly preferable to the baselines in terms of identity preservation, the quality of the produced images, and the identification of the essential attributes of an image. 

\subsubsection{Acknowledgment}
This project has received funding from the European Research Council (ERC) under the European Union's Horizon 2020 research and innovation programme (grant ERC CoG 725974). The first author is further supported by the Council for Higher Education in Israel. The contribution of the first author is part of a PhD thesis research conducted at Tel Aviv University.


\clearpage
\setcounter{equation}{0}
\setcounter{figure}{0}

\title{Image-based CLIP-Guided Essence Transfer} 

\titlerunning{Image-based CLIP-Guided Essence Transfer: Supplementary Material}
\authorrunning{H. Chefer et al.} 
\author{Supplementary Material}
\institute{}

\maketitle

\section{Qualitative Results}
\begin{figure}[h]
    \begin{center}
    \renewcommand{\arraystretch}{0.2}
\begin{tabular}{c@{~~~}cc@{~}c@{~}c@{~}c@{~}c@{~}c@{~}c@{~}c}
\centering{\begin{small}  Target \end{small}}
&
\centering{\begin{small} \begin{turn}{90}  ~~Source \end{turn} \end{small}}
&
\includegraphics[width=0.11\linewidth,clip] {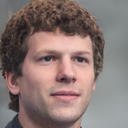}&
\includegraphics[width=0.11\linewidth,clip] {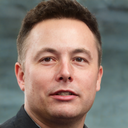}&
\includegraphics[width=0.11\linewidth,clip] {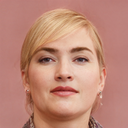}&
\includegraphics[width=0.11\linewidth,clip] {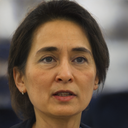}&
\includegraphics[width=0.11\linewidth,clip] {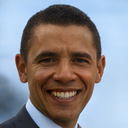}&
\includegraphics[width=0.11\linewidth,clip] {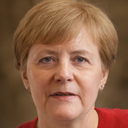}&
\includegraphics[width=0.11\linewidth,clip] {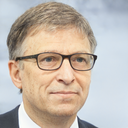}&
\\
\includegraphics[width=0.11\linewidth, clip]{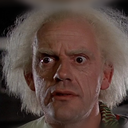}  
&
&
\includegraphics[width=0.11\linewidth,clip] {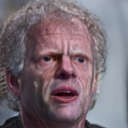}&
\includegraphics[width=0.11\linewidth,clip] {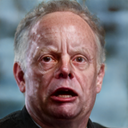}&
\includegraphics[width=0.11\linewidth,clip] {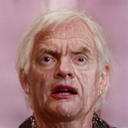}&
\includegraphics[width=0.11\linewidth,clip] {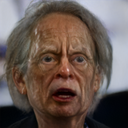}&
\includegraphics[width=0.11\linewidth,clip] {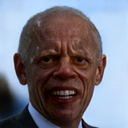}&
\includegraphics[width=0.11\linewidth,clip] {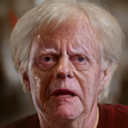}&
\includegraphics[width=0.11\linewidth,clip] {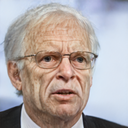}&
\\
\includegraphics[width=0.11\linewidth, clip]{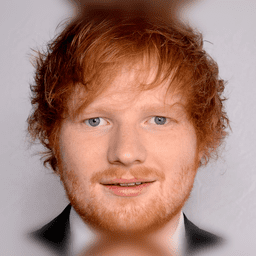}  
&
&
\includegraphics[width=0.11\linewidth,clip] {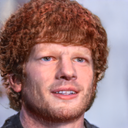} 
&
\includegraphics[width=0.11\linewidth,clip] {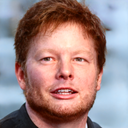}  &
\includegraphics[width=0.11\linewidth, clip]{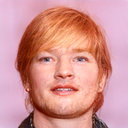}  
&
\includegraphics[width=0.11\linewidth,clip] {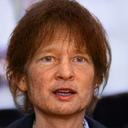} 
&
\includegraphics[width=0.11\linewidth,clip] {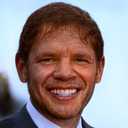}  &
\includegraphics[width=0.11\linewidth,clip] {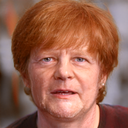}  &
\includegraphics[width=0.11\linewidth,clip] {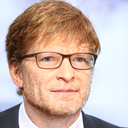}  &
\\
\includegraphics[width=0.11\linewidth, clip]{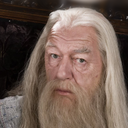} &
&
\includegraphics[width=0.11\linewidth,clip] {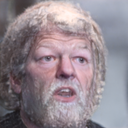} 
&
\includegraphics[width=0.11\linewidth,clip] {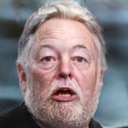}  &
\includegraphics[width=0.11\linewidth, clip]{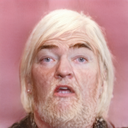}  
&
\includegraphics[width=0.11\linewidth,clip] {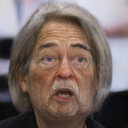} 
&
\includegraphics[width=0.11\linewidth,clip] {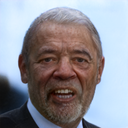}  &
\includegraphics[width=0.11\linewidth,clip] {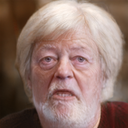}  &
\includegraphics[width=0.11\linewidth,clip] {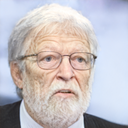}  &\\
\includegraphics[width=0.11\linewidth, clip]{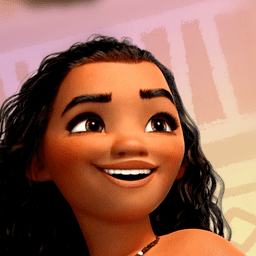}  
&
&
\includegraphics[width=0.11\linewidth,clip] {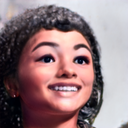} 
&
\includegraphics[width=0.11\linewidth,clip] {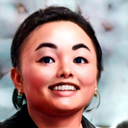}  &
\includegraphics[width=0.11\linewidth, clip]{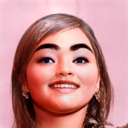}  
&
\includegraphics[width=0.11\linewidth,clip] {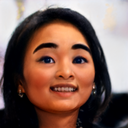} 
&
\includegraphics[width=0.11\linewidth,clip] {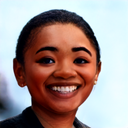}  &
\includegraphics[width=0.11\linewidth,clip] {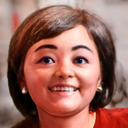}  &
\includegraphics[width=0.11\linewidth,clip] {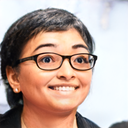}  &\\
\end{tabular}
    \captionof{figure}{
    Examples of using our encoder with the same sources and targets as in Fig. 3 in the main paper. The manipulations are consistent and induce the same semantic manipulation for a wide range of diverse sources.
    }
    \label{fig:encoder_supp}
\end{center}
    \end{figure}

\begin{figure}[h]
    \begin{center}
    \renewcommand{\arraystretch}{0.2}
\begin{tabular}{c@{~~~}cc@{~}c@{~}c@{~}c@{~}c@{~}c@{~}c@{~}c}
\centering{\begin{small}  Target \end{small}}
&
\centering{\begin{small}\begin{turn}{90}  ~Source \end{turn} \end{small}}
&
\includegraphics[width=0.11\linewidth,clip] {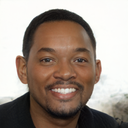}&
\includegraphics[width=0.11\linewidth,clip] {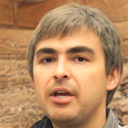}&
\includegraphics[width=0.11\linewidth,clip] {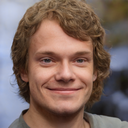}&
\includegraphics[width=0.11\linewidth,clip] {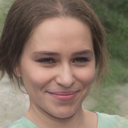}&
\includegraphics[width=0.11\linewidth,clip] {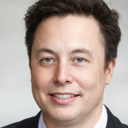}&
\includegraphics[width=0.11\linewidth,clip] {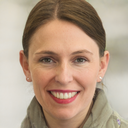}&
\includegraphics[width=0.11\linewidth,clip] {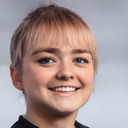}&
\\
\includegraphics[width=0.11\linewidth, clip]{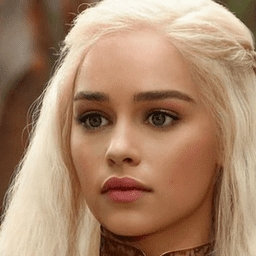}
&
&
\includegraphics[width=0.11\linewidth,clip] {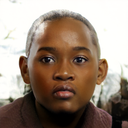}&
\includegraphics[width=0.11\linewidth,clip] {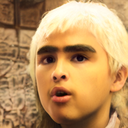}&
\includegraphics[width=0.11\linewidth,clip] {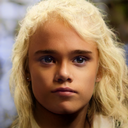}&
\includegraphics[width=0.11\linewidth,clip] {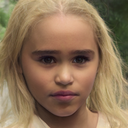}&
\includegraphics[width=0.11\linewidth,clip] {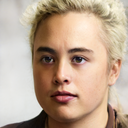}&
\includegraphics[width=0.11\linewidth,clip] {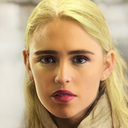}&
\includegraphics[width=0.11\linewidth,clip] {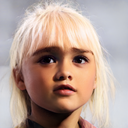}&
\\
\includegraphics[width=0.11\linewidth, clip]{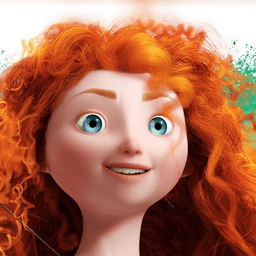}  
&
&
\includegraphics[width=0.11\linewidth,clip] {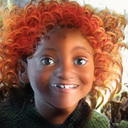}&
\includegraphics[width=0.11\linewidth,clip] {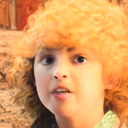}&
\includegraphics[width=0.11\linewidth,clip] {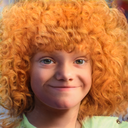}&
\includegraphics[width=0.11\linewidth,clip] {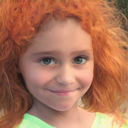}&
\includegraphics[width=0.11\linewidth,clip] {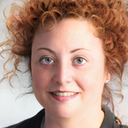}&
\includegraphics[width=0.11\linewidth,clip] {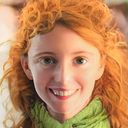}&
\includegraphics[width=0.11\linewidth,clip] {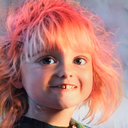}&
\\
\includegraphics[width=0.11\linewidth, clip]{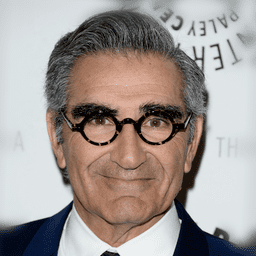}  
&
&
\includegraphics[width=0.11\linewidth,clip] {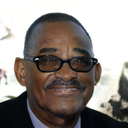}&
\includegraphics[width=0.11\linewidth,clip] {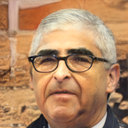}&
\includegraphics[width=0.11\linewidth,clip] {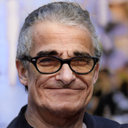}&
\includegraphics[width=0.11\linewidth,clip] {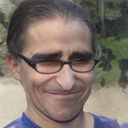}&
\includegraphics[width=0.11\linewidth,clip] {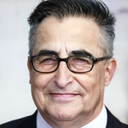}&
\includegraphics[width=0.11\linewidth,clip] {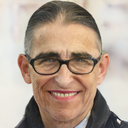}&
\includegraphics[width=0.11\linewidth,clip] {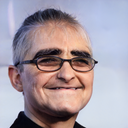}&\\
\includegraphics[width=0.11\linewidth, clip]{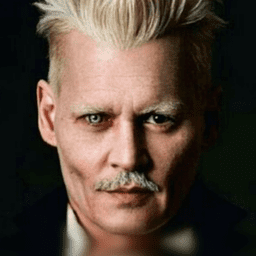}  
&
&
\includegraphics[width=0.11\linewidth,clip] {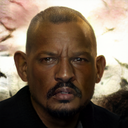}&
\includegraphics[width=0.11\linewidth,clip] {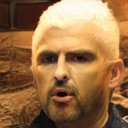}&
\includegraphics[width=0.11\linewidth,clip] {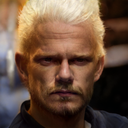}&
\includegraphics[width=0.11\linewidth,clip] {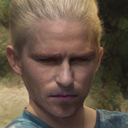}&
\includegraphics[width=0.11\linewidth,clip] {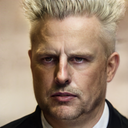}&
\includegraphics[width=0.11\linewidth,clip] {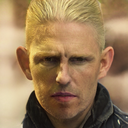}&
\includegraphics[width=0.11\linewidth,clip] {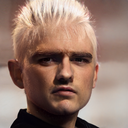}&\\
~\\
\end{tabular}
    \captionof{figure}{Examples of using our optimizer with the same sources and targets as in Fig. 4 in the main paper. The manipulations are consistent and induce the same semantic manipulation for a wide range of diverse sources.
    }
    \label{fig:optimizer}
\end{center}
    \end{figure}

In this section, we provide the complementary versions of Fig.~3, Fig.~4 from the paper, i.e. we present the sources and targets from Fig.~3 manipulated with the encoder, and the sources and targets from Fig.~4 manipulated with the optimizer. Fig.~\ref{fig:encoder_supp} complements Fig.~3 from the paper, and Fig.~\ref{fig:optimizer} complements Fig.~4 from the paper. The Dumbledore target in Fig.~\ref{fig:encoder_supp} is slightly different than the one used for the optimizer since the target we used for the optimizer cannot be aligned, which is a precondition for our encoder to process an image. As can be seen in both Fig.~\ref{fig:encoder_supp} and Fig.~\ref{fig:optimizer}, both our methods produce meaningful manipulations, and successfully transfer notable semantic attributes from the targets to the various sources, while preserving the identity of the source images. 

\clearpage

\section{Celebrity Targets}
\begin{figure}[h]
    \begin{center}
    \renewcommand{\arraystretch}{0.2}
\begin{tabular}{c@{~}c@{~}c@{~}c@{~}c}

\includegraphics[width=0.13\linewidth,clip] {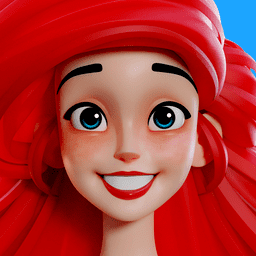}&
\includegraphics[width=0.13\linewidth,clip] {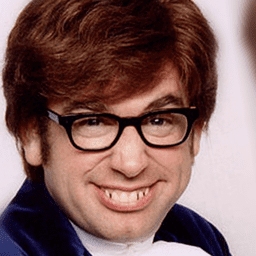}&
\includegraphics[width=0.13\linewidth,clip] {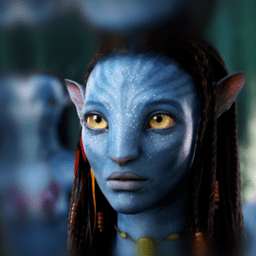}&
\includegraphics[width=0.13\linewidth,clip] {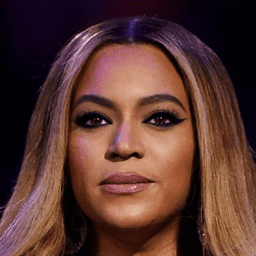}&
\includegraphics[width=0.13\linewidth,clip] {celebrity_targets/brave.png}
\\
\includegraphics[width=0.13\linewidth,clip] {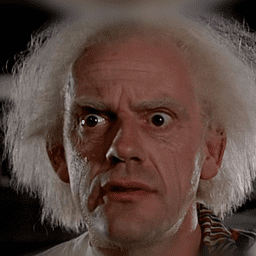}&
\includegraphics[width=0.13\linewidth,clip] {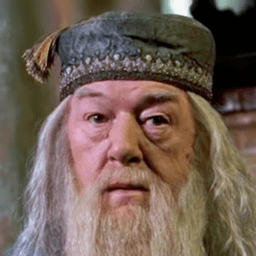}&
\includegraphics[width=0.13\linewidth,clip] {celebrity_targets/ed_sheeran.png}&
\includegraphics[width=0.13\linewidth,clip] {celebrity_targets/Eugene_Levy.png}&
\includegraphics[width=0.13\linewidth,clip] {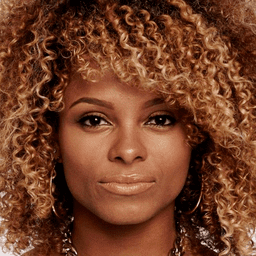}
\\
\includegraphics[width=0.13\linewidth,clip] {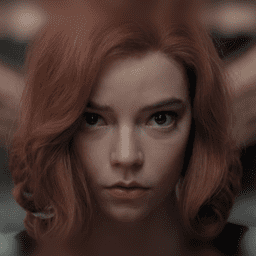}&
\includegraphics[width=0.13\linewidth,clip] {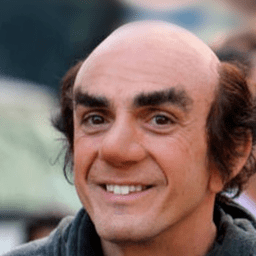}&
\includegraphics[width=0.13\linewidth,clip] {celebrity_targets/grindelwald.png}&
\includegraphics[width=0.13\linewidth,clip] {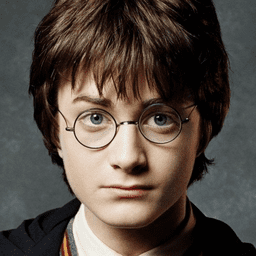}&
\includegraphics[width=0.13\linewidth,clip] {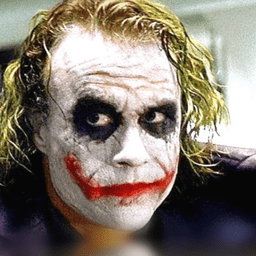}
\\
\includegraphics[width=0.13\linewidth,clip] {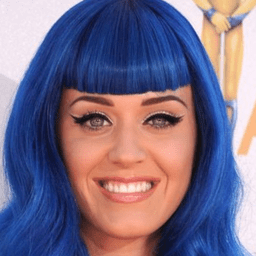}&
\includegraphics[width=0.13\linewidth,clip] {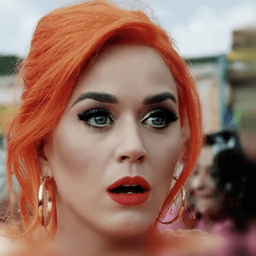}&
\includegraphics[width=0.13\linewidth,clip] {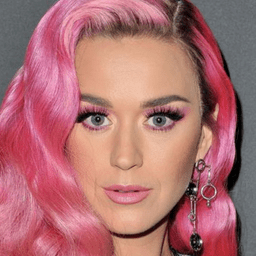}&
\includegraphics[width=0.13\linewidth,clip] {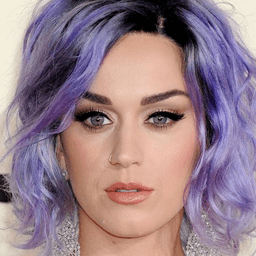}&
\includegraphics[width=0.13\linewidth,clip] {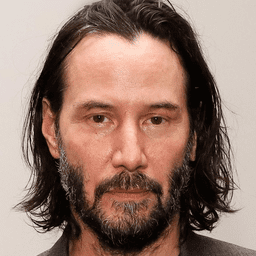}
\\
\includegraphics[width=0.13\linewidth,clip] {celebrity_targets/khaleesi.png}&
\includegraphics[width=0.13\linewidth,clip] {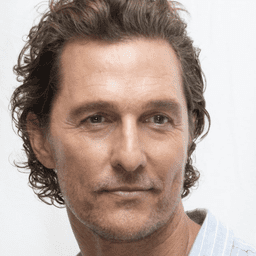}&
\includegraphics[width=0.13\linewidth,clip] {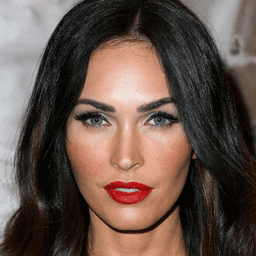}&
\includegraphics[width=0.13\linewidth,clip] {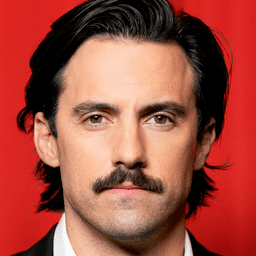}&
\includegraphics[width=0.13\linewidth,clip] {celebrity_targets/moana.png}
\\
\includegraphics[width=0.13\linewidth,clip] {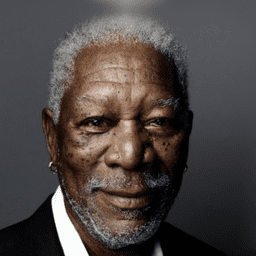}&
\includegraphics[width=0.13\linewidth,clip] {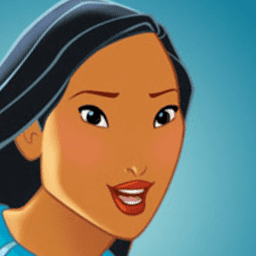}&
\includegraphics[width=0.13\linewidth,clip] {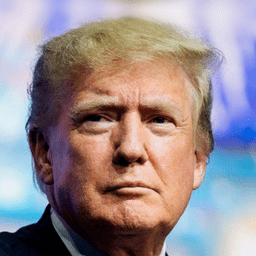}&
\includegraphics[width=0.13\linewidth,clip] {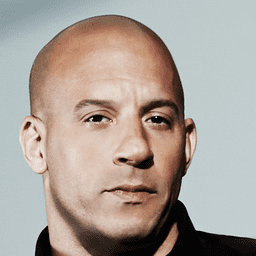}&
\includegraphics[width=0.13\linewidth,clip] {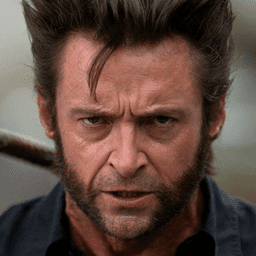}
\\
&
&
\includegraphics[width=0.13\linewidth,clip] {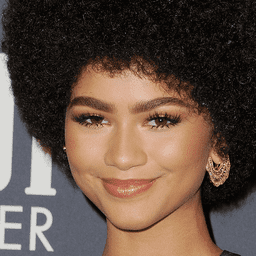}&
&
\\
\end{tabular}
    \captionof{figure}{Well-known characters used as targets for our celebrities test. The chosen targets are diverse in terms of gender, ethnicity, and the attributes that they demonstrate.
    }
    \label{fig:celeb_targets}
\end{center}
    \end{figure}

\
Fig.~\ref{fig:celeb_targets} presents all the targets used in our celebrities test. As can be seen, our targets demonstrate notable or extreme semantic properties, such as unusual hair colors and styles, beards, glasses, as well as a variety of ages, genders, and ethnicities, and also contain out-of-domain animated characters.

\clearpage

\section{Qualitative Comparisons}
\begin{figure}[h]
    \begin{center}
    \renewcommand{\arraystretch}{0.2}
\begin{tabular}{c@{~~}ccccc@{~~~~~~}cccc}
\centering{  Target }
&
\centering{\begin{turn}{90}~~ Src. \end{turn} }
&
\includegraphics[width=0.085\linewidth,clip] {overfit/3.png}&
\includegraphics[width=0.085\linewidth,clip] {overfit/15.png}&
\includegraphics[width=0.085\linewidth,clip] {overfit/44.png}&
\includegraphics[width=0.085\linewidth,clip] {overfit/62.png}&
\includegraphics[width=0.085\linewidth,clip] {overfit/3.png}&
\includegraphics[width=0.085\linewidth,clip] {overfit/15.png}&
\includegraphics[width=0.085\linewidth,clip] {overfit/44.png}&
\includegraphics[width=0.085\linewidth,clip] {overfit/62.png}
\\
\includegraphics[width=0.085\linewidth, clip]{overfit/ariel/target.png}
&
&
\includegraphics[width=0.085\linewidth,clip] {overfit/ariel/ours/3.png} 
&
\includegraphics[width=0.085\linewidth,clip] {overfit/ariel/ours/15.png} &
\includegraphics[width=0.085\linewidth, clip]{overfit/ariel/ours/44.png}  
&
\includegraphics[width=0.085\linewidth,clip] {overfit/ariel/ours/62.png} 
&
\includegraphics[width=0.085\linewidth,clip] {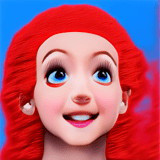} 
&
\includegraphics[width=0.085\linewidth,clip] {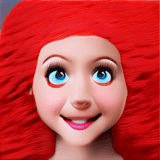} &
\includegraphics[width=0.085\linewidth, clip]{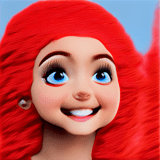}  
&
\includegraphics[width=0.085\linewidth,clip] {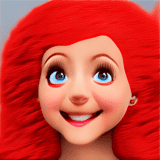} 
\\
\includegraphics[width=0.085\linewidth, clip]{overfit/dumbledore/target.png}
&
&
\includegraphics[width=0.085\linewidth,clip] {overfit/dumbledore/ours/3.png} 
&
\includegraphics[width=0.085\linewidth,clip] {overfit/dumbledore/ours/15.png} &
\includegraphics[width=0.085\linewidth, clip]{overfit/dumbledore/ours/44.png}  
&
\includegraphics[width=0.085\linewidth,clip] {overfit/dumbledore/ours/62.png} 
&
\includegraphics[width=0.085\linewidth,clip] {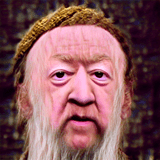} 
&
\includegraphics[width=0.085\linewidth,clip] {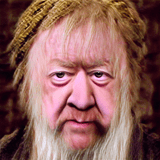} &
\includegraphics[width=0.085\linewidth, clip]{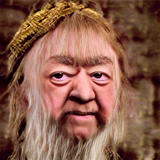}  
&
\includegraphics[width=0.085\linewidth,clip] {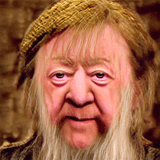} 
\\
\includegraphics[width=0.085\linewidth, clip]{overfit/wolverine/target.png}
&
&
\includegraphics[width=0.085\linewidth,clip] {overfit/wolverine/ours/3.png} 
&
\includegraphics[width=0.085\linewidth,clip] {overfit/wolverine/ours/15.png} &
\includegraphics[width=0.085\linewidth, clip]{overfit/wolverine/ours/44.png}  
&
\includegraphics[width=0.085\linewidth,clip] {overfit/wolverine/ours/62.png} 
&
\includegraphics[width=0.085\linewidth,clip] {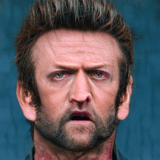} 
&
\includegraphics[width=0.085\linewidth,clip] {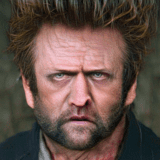} &
\includegraphics[width=0.085\linewidth, clip]{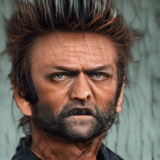}  
&
\includegraphics[width=0.085\linewidth,clip] {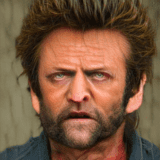} 
\\
\includegraphics[width=0.085\linewidth, clip]{overfit/keanu/target.png}
&
&
\includegraphics[width=0.085\linewidth,clip] {overfit/keanu/ours/3.png} 
&
\includegraphics[width=0.085\linewidth,clip] {overfit/keanu/ours/15.png} &
\includegraphics[width=0.085\linewidth, clip]{overfit/keanu/ours/44.png}  
&
\includegraphics[width=0.085\linewidth,clip] {overfit/keanu/ours/62.png} 
&
\includegraphics[width=0.085\linewidth,clip] {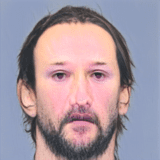} 
&
\includegraphics[width=0.085\linewidth,clip] {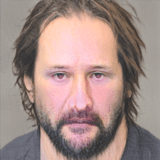} &
\includegraphics[width=0.085\linewidth, clip]{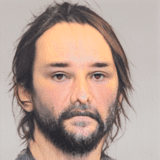}  
&
\includegraphics[width=0.085\linewidth,clip] {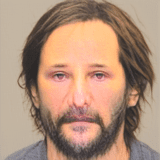} 
\\
\includegraphics[width=0.085\linewidth, clip]{overfit/ed_sheeran/target.png}
&
&
\includegraphics[width=0.085\linewidth,clip] {overfit/ed_sheeran/ours/3.png} 
&
\includegraphics[width=0.085\linewidth,clip] {overfit/ed_sheeran/ours/15.png} &
\includegraphics[width=0.085\linewidth, clip]{overfit/ed_sheeran/ours/44.png}  
&
\includegraphics[width=0.085\linewidth,clip] {overfit/ed_sheeran/ours/62.png} 
&
\includegraphics[width=0.085\linewidth,clip] {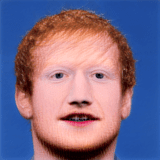} 
&
\includegraphics[width=0.085\linewidth,clip] {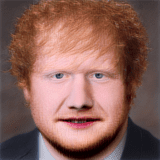} &
\includegraphics[width=0.085\linewidth, clip]{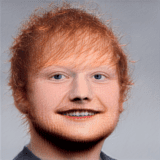}  
&
\includegraphics[width=0.085\linewidth,clip] {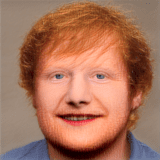} 
\\
\includegraphics[width=0.085\linewidth, clip]{overfit/potter/target.png}
&
&
\includegraphics[width=0.085\linewidth,clip] {overfit/potter/ours/3.png} 
&
\includegraphics[width=0.085\linewidth,clip] {overfit/potter/ours/15.png} &
\includegraphics[width=0.085\linewidth, clip]{overfit/potter/ours/44.png}  
&
\includegraphics[width=0.085\linewidth,clip] {overfit/potter/ours/62.png} 
&
\includegraphics[width=0.085\linewidth,clip] {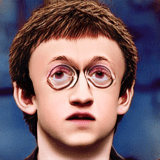} 
&
\includegraphics[width=0.085\linewidth,clip] {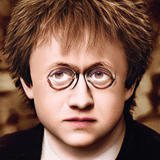} &
\includegraphics[width=0.085\linewidth, clip]{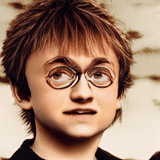}  
&
\includegraphics[width=0.085\linewidth,clip] {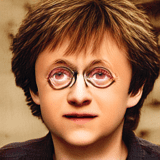} \\
 & & & \multicolumn{2}{c}{ Ours} & & &  \multicolumn{2}{c}{NADA}\\
\end{tabular}
    \captionof{figure}{Comparison to StyleGAN-NADA from our celebrities test. First three rows are manipulations with our optimizer, and last three with our encoder.
    }
    \label{fig:nada}
\end{center}
    \end{figure}

\
In this section, we enclose the qualitative comparison to StyleGAN-NADA that was omitted from Fig.~6 of the main text (see Fig.~\ref{fig:nada}), as well as qualitative comparisons from our FFHQ experiments, which do not appear in the main text. 

As can be seen from Fig.~\ref{fig:nada}, StyleGAN-NADA falls short on preserving the source identity, and results in either a uniform identity or a mixed identity of the source and the target, as mentioned in the main text. 

\begin{figure}[h]
    \begin{center}
    \renewcommand{\arraystretch}{0.2}
\begin{tabular}{c@{~~}ccccc@{~~}cccc@{~~}cccc}
\centering{  Target }
&
\centering{\begin{turn}{90}~ Src. \end{turn} }
&
\includegraphics[width=0.065\linewidth,clip] {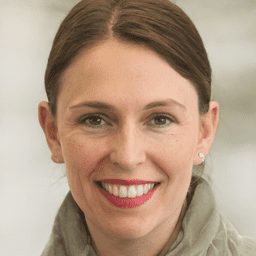}&
\includegraphics[width=0.065\linewidth,clip] {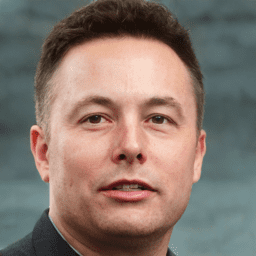}&
\includegraphics[width=0.065\linewidth,clip] {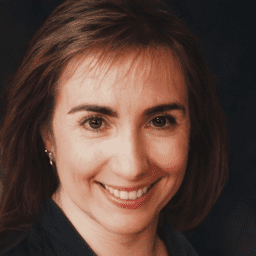}&
\includegraphics[width=0.065\linewidth,clip] {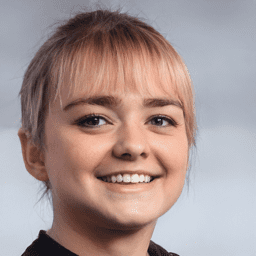}&
\includegraphics[width=0.065\linewidth,clip] {overfit_supp/2.png}&
\includegraphics[width=0.065\linewidth,clip] {overfit_supp/12.png}&
\includegraphics[width=0.065\linewidth,clip] {overfit_supp/24.png}&
\includegraphics[width=0.065\linewidth,clip] {overfit_supp/44.png}&
\includegraphics[width=0.065\linewidth,clip] {overfit_supp/2.png}&
\includegraphics[width=0.065\linewidth,clip] {overfit_supp/12.png}&
\includegraphics[width=0.065\linewidth,clip] {overfit_supp/24.png}&
\includegraphics[width=0.065\linewidth,clip] {overfit_supp/44.png}
\\
\includegraphics[width=0.065\linewidth,clip] {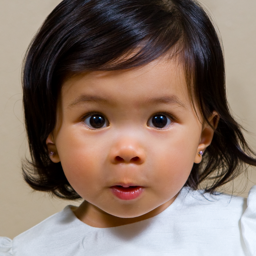}&
&
\includegraphics[width=0.065\linewidth,clip] {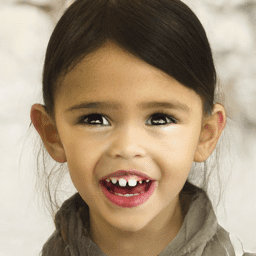}&
\includegraphics[width=0.065\linewidth,clip] {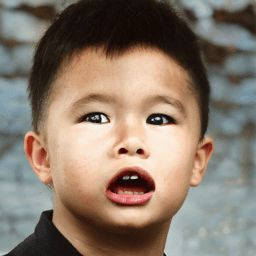}&
\includegraphics[width=0.065\linewidth,clip] {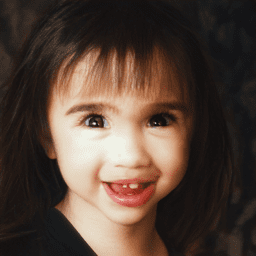}&
\includegraphics[width=0.065\linewidth,clip] {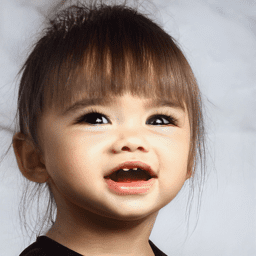}&
\includegraphics[width=0.065\linewidth,clip] {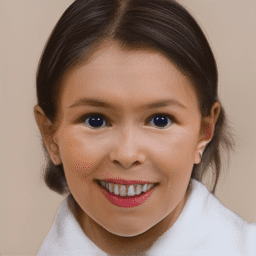}&
\includegraphics[width=0.065\linewidth,clip] {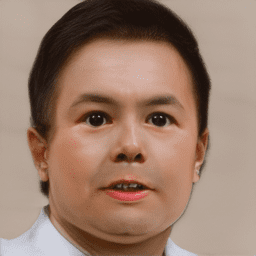}&
\includegraphics[width=0.065\linewidth,clip] {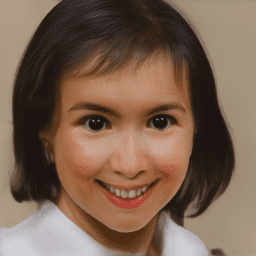}&
\includegraphics[width=0.065\linewidth,clip] {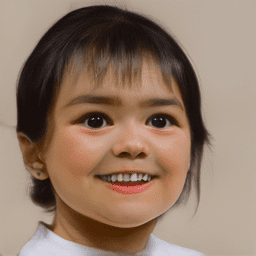}&
\includegraphics[width=0.065\linewidth,clip] {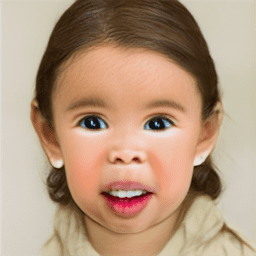}&
\includegraphics[width=0.065\linewidth,clip] {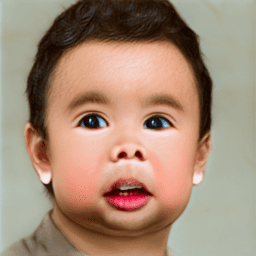}&
\includegraphics[width=0.065\linewidth,clip] {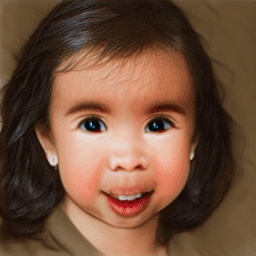}&
\includegraphics[width=0.065\linewidth,clip] {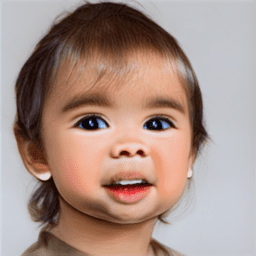}
\\
\includegraphics[width=0.065\linewidth,clip] {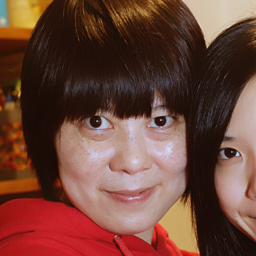}&
&
\includegraphics[width=0.065\linewidth,clip] {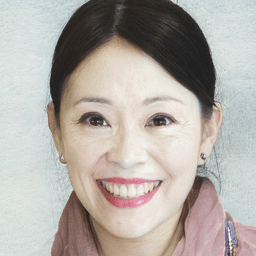}&
\includegraphics[width=0.065\linewidth,clip] {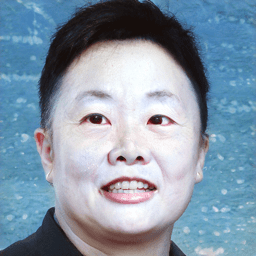}&
\includegraphics[width=0.065\linewidth,clip] {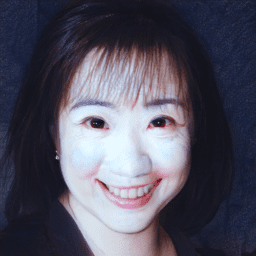}&
\includegraphics[width=0.065\linewidth,clip] {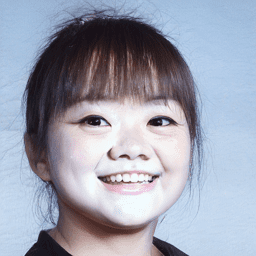}&
\includegraphics[width=0.065\linewidth,clip] {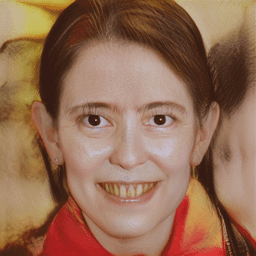}&
\includegraphics[width=0.065\linewidth,clip] {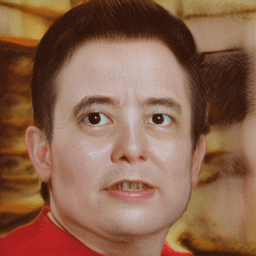}&
\includegraphics[width=0.065\linewidth,clip] {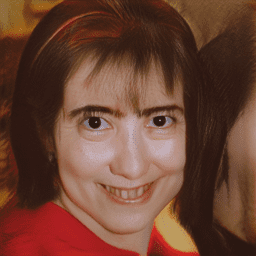}&
\includegraphics[width=0.065\linewidth,clip] {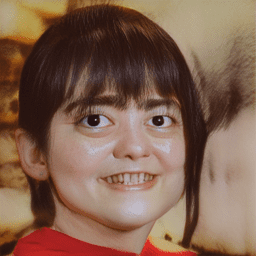}&
\includegraphics[width=0.065\linewidth,clip] {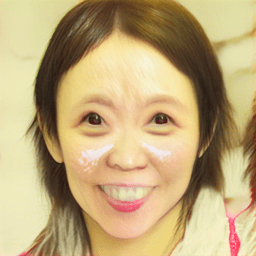}&
\includegraphics[width=0.065\linewidth,clip] {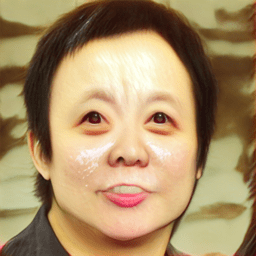}&
\includegraphics[width=0.065\linewidth,clip] {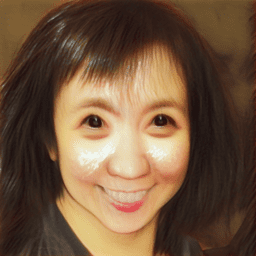}&
\includegraphics[width=0.065\linewidth,clip] {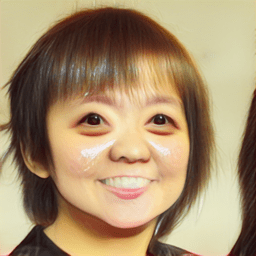}
\\
\includegraphics[width=0.065\linewidth,clip] {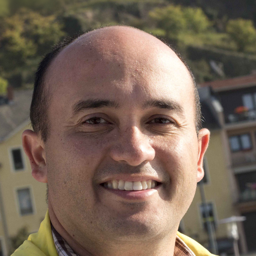}&
&
\includegraphics[width=0.065\linewidth,clip] {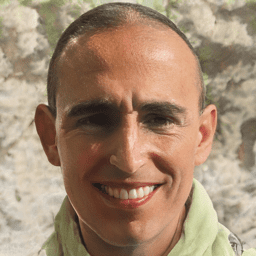}&
\includegraphics[width=0.065\linewidth,clip] {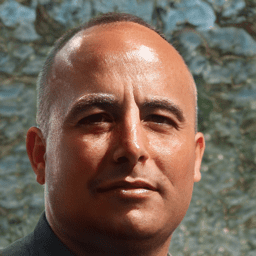}&
\includegraphics[width=0.065\linewidth,clip] {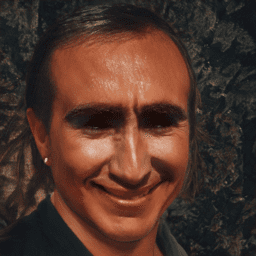}&
\includegraphics[width=0.065\linewidth,clip] {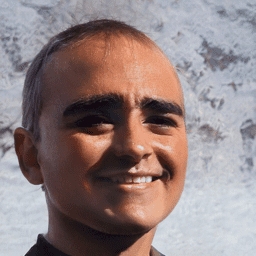}&
\includegraphics[width=0.065\linewidth,clip] {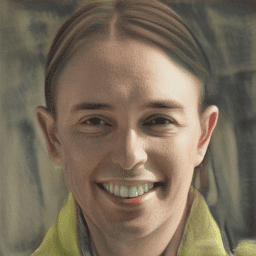}&
\includegraphics[width=0.065\linewidth,clip] {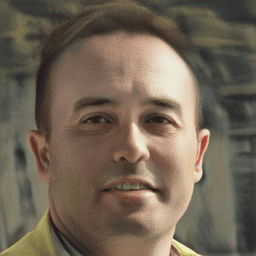}&
\includegraphics[width=0.065\linewidth,clip] {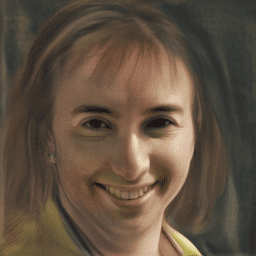}&
\includegraphics[width=0.065\linewidth,clip] {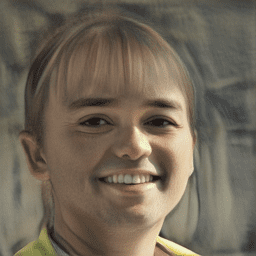}&
\includegraphics[width=0.065\linewidth,clip] {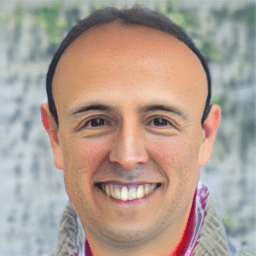}&
\includegraphics[width=0.065\linewidth,clip] {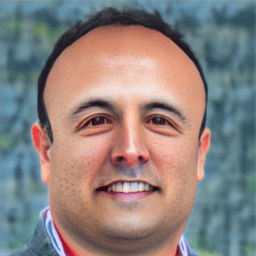}&
\includegraphics[width=0.065\linewidth,clip] {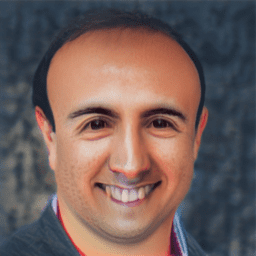}&
\includegraphics[width=0.065\linewidth,clip] {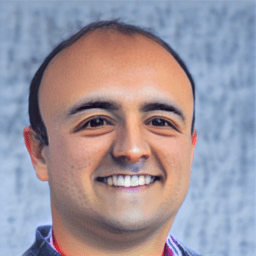}
\\
\includegraphics[width=0.065\linewidth,clip] {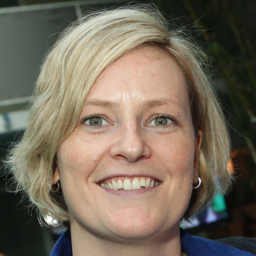}&
&
\includegraphics[width=0.065\linewidth,clip] {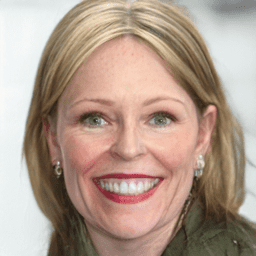}&
\includegraphics[width=0.065\linewidth,clip] {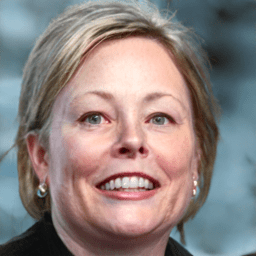}&
\includegraphics[width=0.065\linewidth,clip] {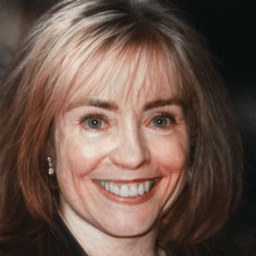}&
\includegraphics[width=0.065\linewidth,clip] {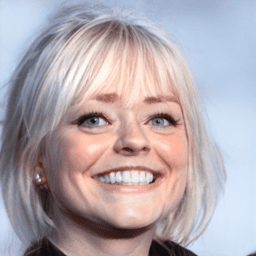}&
\includegraphics[width=0.065\linewidth,clip] {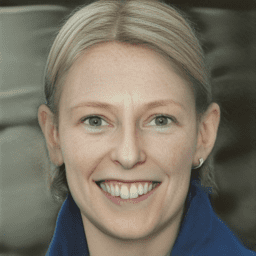}&
\includegraphics[width=0.065\linewidth,clip] {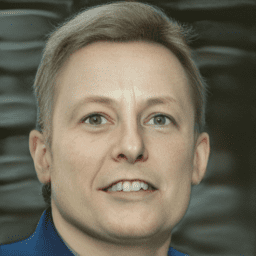}&
\includegraphics[width=0.065\linewidth,clip] {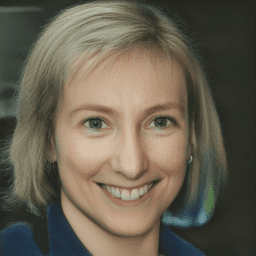}&
\includegraphics[width=0.065\linewidth,clip] {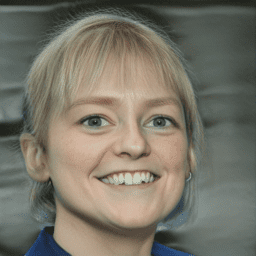}&
\includegraphics[width=0.065\linewidth,clip] {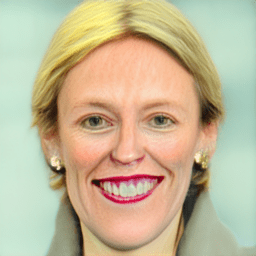}&
\includegraphics[width=0.065\linewidth,clip] {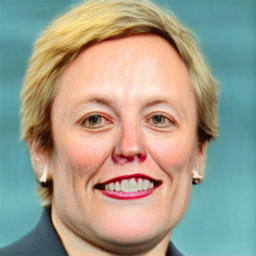}&
\includegraphics[width=0.065\linewidth,clip] {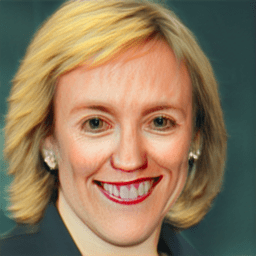}&
\includegraphics[width=0.065\linewidth,clip] {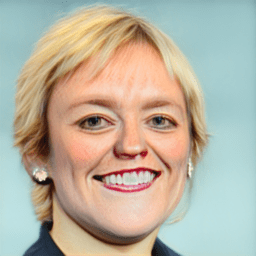}
\\
\includegraphics[width=0.065\linewidth,clip] {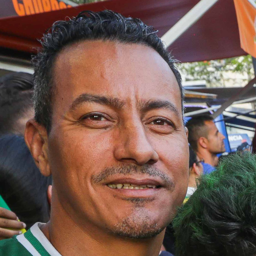}&
&
\includegraphics[width=0.065\linewidth,clip] {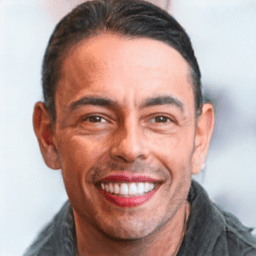}&
\includegraphics[width=0.065\linewidth,clip] {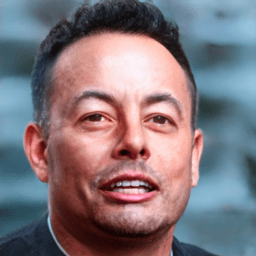}&
\includegraphics[width=0.065\linewidth,clip] {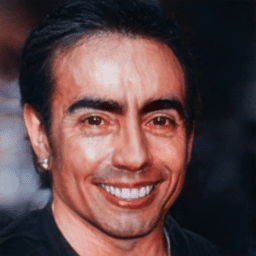}&
\includegraphics[width=0.065\linewidth,clip] {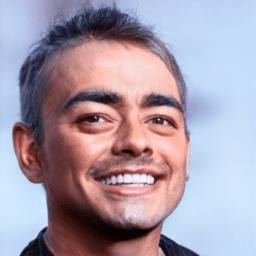}&
\includegraphics[width=0.065\linewidth,clip] {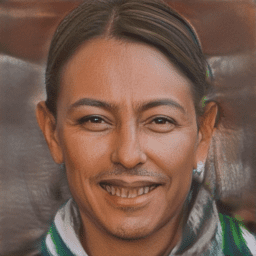}&
\includegraphics[width=0.065\linewidth,clip] {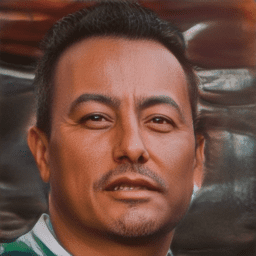}&
\includegraphics[width=0.065\linewidth,clip] {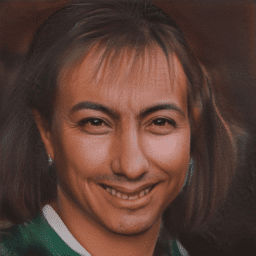}&
\includegraphics[width=0.065\linewidth,clip] {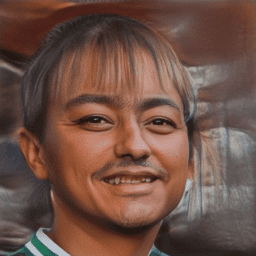}&
\includegraphics[width=0.065\linewidth,clip] {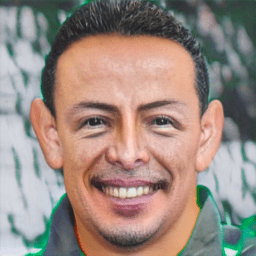}&
\includegraphics[width=0.065\linewidth,clip] {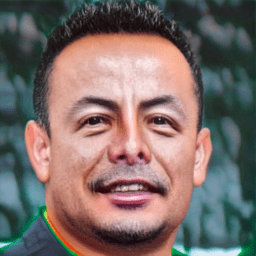}&
\includegraphics[width=0.065\linewidth,clip] {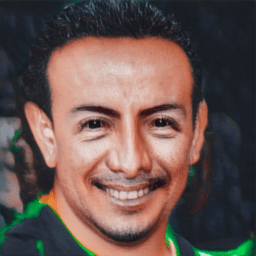}&
\includegraphics[width=0.065\linewidth,clip] {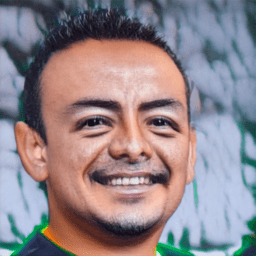}
\\
\includegraphics[width=0.065\linewidth,clip] {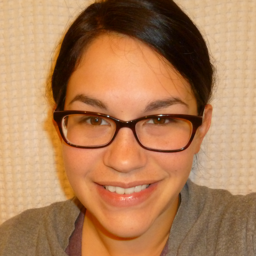}&
&
\includegraphics[width=0.065\linewidth,clip] {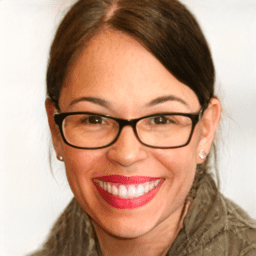}&
\includegraphics[width=0.065\linewidth,clip] {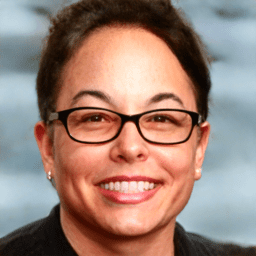}&
\includegraphics[width=0.065\linewidth,clip] {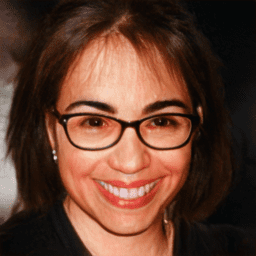}&
\includegraphics[width=0.065\linewidth,clip] {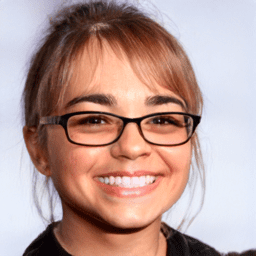}&
\includegraphics[width=0.065\linewidth,clip] {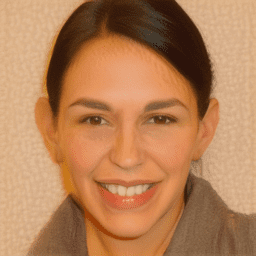}&
\includegraphics[width=0.065\linewidth,clip] {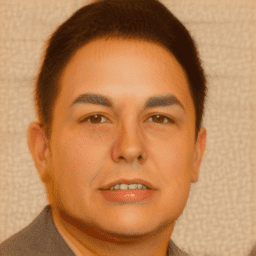}&
\includegraphics[width=0.065\linewidth,clip] {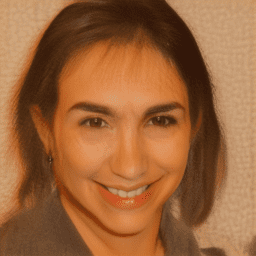}&
\includegraphics[width=0.065\linewidth,clip] {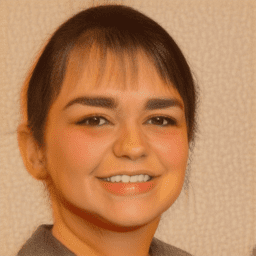}&
\includegraphics[width=0.065\linewidth,clip] {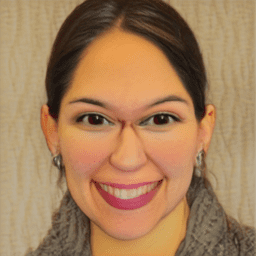}&
\includegraphics[width=0.065\linewidth,clip] {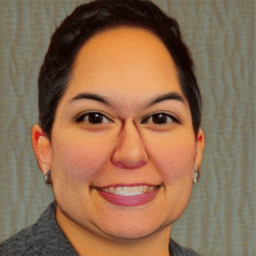}&
\includegraphics[width=0.065\linewidth,clip] {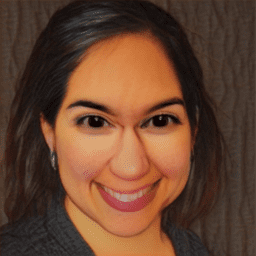}&
\includegraphics[width=0.065\linewidth,clip] {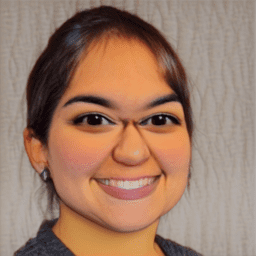}
\\\\
 & & & \multicolumn{2}{c}{Ours} & & &   \multicolumn{2}{c}{JoJoGAN}& & & \multicolumn{2}{c}{NADA}\\
\end{tabular}
    \captionof{figure}{Comparison to methods that suffer from a high loss of source identity on our FFHQ test. First three rows are manipulations with our optimizer, and last three with our encoder. 
    }
    \label{fig:overfit_supp}
\end{center}
    \end{figure}

\begin{figure}[h]
    \begin{center}
    \renewcommand{\arraystretch}{0.2}
\begin{tabular}{c@{~~}ccccc@{~~~~~~}cccc}
\centering{  Target }
&
\centering{\begin{turn}{90}~~ Src. \end{turn} }
&
\includegraphics[width=0.085\linewidth,clip] {overfit_supp/2.png}&
\includegraphics[width=0.085\linewidth,clip] {overfit_supp/12.png}&
\includegraphics[width=0.085\linewidth,clip] {overfit_supp/24.png}&
\includegraphics[width=0.085\linewidth,clip] {overfit_supp/44.png}&
\includegraphics[width=0.085\linewidth,clip] {overfit_supp/2.png}&
\includegraphics[width=0.085\linewidth,clip] {overfit_supp/12.png}&
\includegraphics[width=0.085\linewidth,clip] {overfit_supp/24.png}&
\includegraphics[width=0.085\linewidth,clip] {overfit_supp/44.png}
\\
\includegraphics[width=0.085\linewidth, clip]{overfit_supp/00003/target.png}
&
&
\includegraphics[width=0.085\linewidth,clip] {overfit_supp/00003/ours/2.png} 
&
\includegraphics[width=0.085\linewidth,clip] {overfit_supp/00003/ours/12.png} &
\includegraphics[width=0.085\linewidth, clip]{overfit_supp/00003/ours/24.png}  
&
\includegraphics[width=0.085\linewidth,clip] {overfit_supp/00003/ours/44.png}  
&
\includegraphics[width=0.085\linewidth,clip] {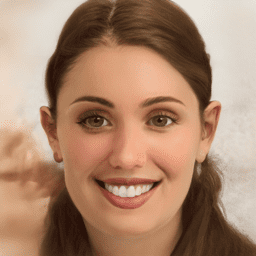} 
&
\includegraphics[width=0.085\linewidth,clip] {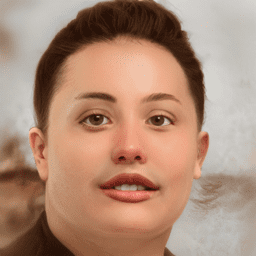} &
\includegraphics[width=0.085\linewidth, clip]{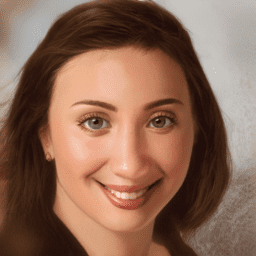}  
&
\includegraphics[width=0.085\linewidth,clip] {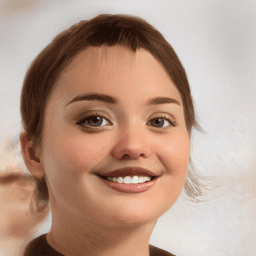}
\\
\includegraphics[width=0.085\linewidth, clip]{overfit_supp/00008/target.png}
&
&
\includegraphics[width=0.085\linewidth,clip] {overfit_supp/00008/ours/2.png} 
&
\includegraphics[width=0.085\linewidth,clip] {overfit_supp/00008/ours/12.png} &
\includegraphics[width=0.085\linewidth, clip]{overfit_supp/00008/ours/24.png}  
&
\includegraphics[width=0.085\linewidth,clip] {overfit_supp/00008/ours/44.png} 
&
\includegraphics[width=0.085\linewidth,clip] {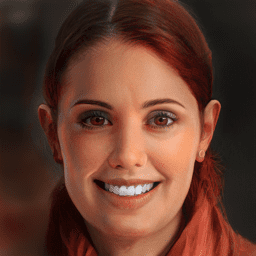} 
&
\includegraphics[width=0.085\linewidth,clip] {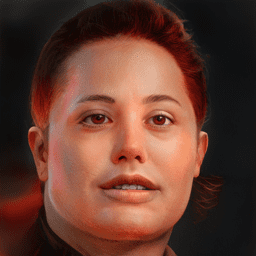} &
\includegraphics[width=0.085\linewidth, clip]{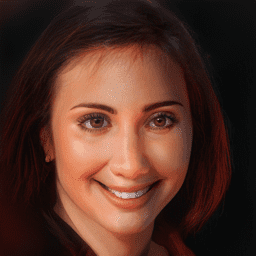}  
&
\includegraphics[width=0.085\linewidth,clip] {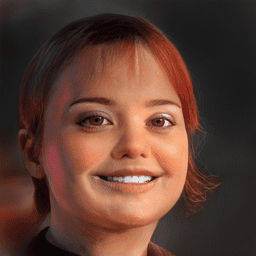} 
\\
\includegraphics[width=0.085\linewidth, clip]{overfit_supp/00045/target.png}
&
&
\includegraphics[width=0.085\linewidth,clip] {overfit_supp/00045/ours/2.png} 
&
\includegraphics[width=0.085\linewidth,clip] {overfit_supp/00045/ours/12.png} &
\includegraphics[width=0.085\linewidth, clip]{overfit_supp/00045/ours/24.png}  
&
\includegraphics[width=0.085\linewidth,clip] {overfit_supp/00045/ours/44.png} 
&
\includegraphics[width=0.085\linewidth,clip] {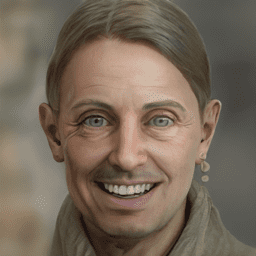} 
&
\includegraphics[width=0.085\linewidth,clip] {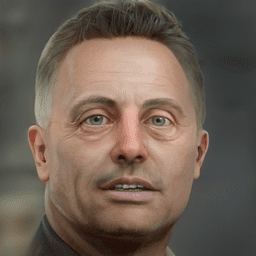} &
\includegraphics[width=0.085\linewidth, clip]{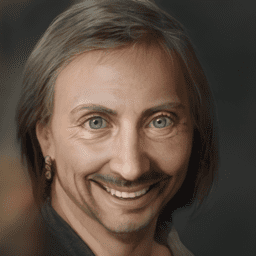}  
&
\includegraphics[width=0.085\linewidth,clip] {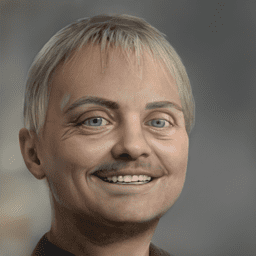}
\\
\includegraphics[width=0.085\linewidth, clip]{overfit_supp/00018/target.png}
&
&
\includegraphics[width=0.085\linewidth,clip] {overfit_supp/00018/ours/2.png} 
&
\includegraphics[width=0.085\linewidth,clip] {overfit_supp/00018/ours/12.png} &
\includegraphics[width=0.085\linewidth, clip]{overfit_supp/00018/ours/24.png}  
&
\includegraphics[width=0.085\linewidth,clip] {overfit_supp/00018/ours/44.png} 
&
\includegraphics[width=0.085\linewidth,clip] {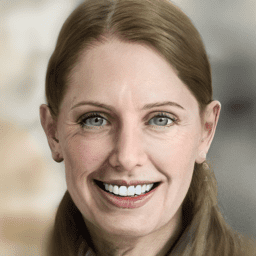} 
&
\includegraphics[width=0.085\linewidth,clip] {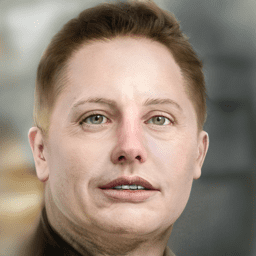} &
\includegraphics[width=0.085\linewidth, clip]{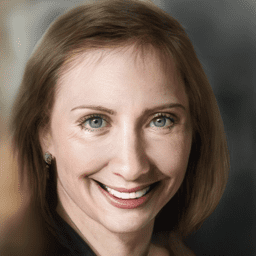}  
&
\includegraphics[width=0.085\linewidth,clip] {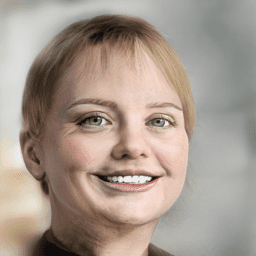}
\\
\includegraphics[width=0.085\linewidth, clip]{overfit_supp/00027/target.png}
&
&
\includegraphics[width=0.085\linewidth,clip] {overfit_supp/00027/ours/2.png} 
&
\includegraphics[width=0.085\linewidth,clip] {overfit_supp/00027/ours/12.png} &
\includegraphics[width=0.085\linewidth, clip]{overfit_supp/00027/ours/24.png}  
&
\includegraphics[width=0.085\linewidth,clip] {overfit_supp/00027/ours/44.png} 
&
\includegraphics[width=0.085\linewidth,clip] {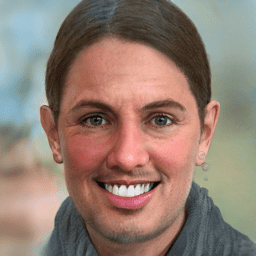} 
&
\includegraphics[width=0.085\linewidth,clip] {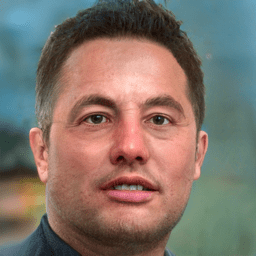} &
\includegraphics[width=0.085\linewidth, clip]{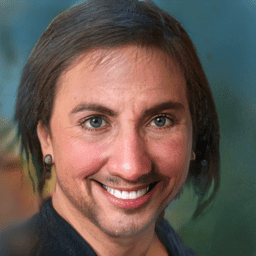}  
&
\includegraphics[width=0.085\linewidth,clip] {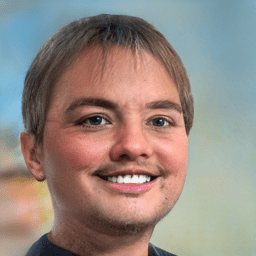} 
\\
\includegraphics[width=0.085\linewidth, clip]{overfit_supp/00037/target.png}
&
&
\includegraphics[width=0.085\linewidth,clip] {overfit_supp/00037/ours/2.png} 
&
\includegraphics[width=0.085\linewidth,clip] {overfit_supp/00037/ours/12.png} &
\includegraphics[width=0.085\linewidth, clip]{overfit_supp/00037/ours/24.png}  
&
\includegraphics[width=0.085\linewidth,clip] {overfit_supp/00037/ours/44.png}  
&
\includegraphics[width=0.085\linewidth,clip] {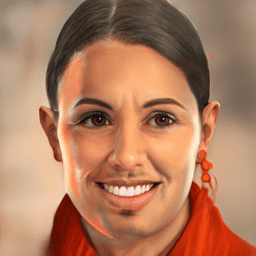} 
&
\includegraphics[width=0.085\linewidth,clip] {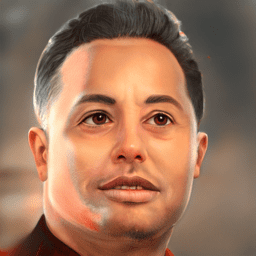} &
\includegraphics[width=0.085\linewidth, clip]{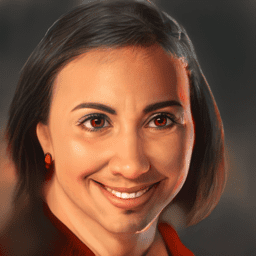}  
&
\includegraphics[width=0.085\linewidth,clip] {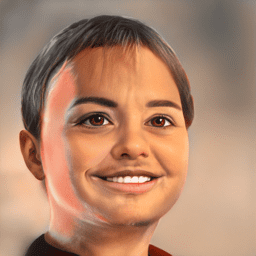}\\
 & & & \multicolumn{2}{c}{ Ours} & & &  \multicolumn{2}{c}{BlendGAN}\\
\end{tabular}
    \captionof{figure}{Comparison to methods that only partially transfer the semantic properties from our FFHQ test. First three rows are manipulations with our optimizer, and the last three
are with our encoder.
    }
    \label{fig:underfit_supp}
\end{center}
    \end{figure}

Fig.~\ref{fig:overfit_supp} presents a comparison between our method and methods that suffer from identity loss on the FFHQ test. The demonstrated results are very similar to those obtained on our celebrities test, where both JoJoGAN and StyleGAN-NADA suffer from a high loss of the source identity. Fig~\ref{fig:underfit_supp} shows a comparison between our method and BlendGAN on the FFHQ test. As in the celebrities test, BlendGAN does not transfer the semantic attributes of the target to the sources.

\clearpage
\section{Essence Decoding}
\begin{figure}[h]
    \begin{center}
    \renewcommand{\arraystretch}{0.2}
\includegraphics[width=0.85\linewidth,clip] {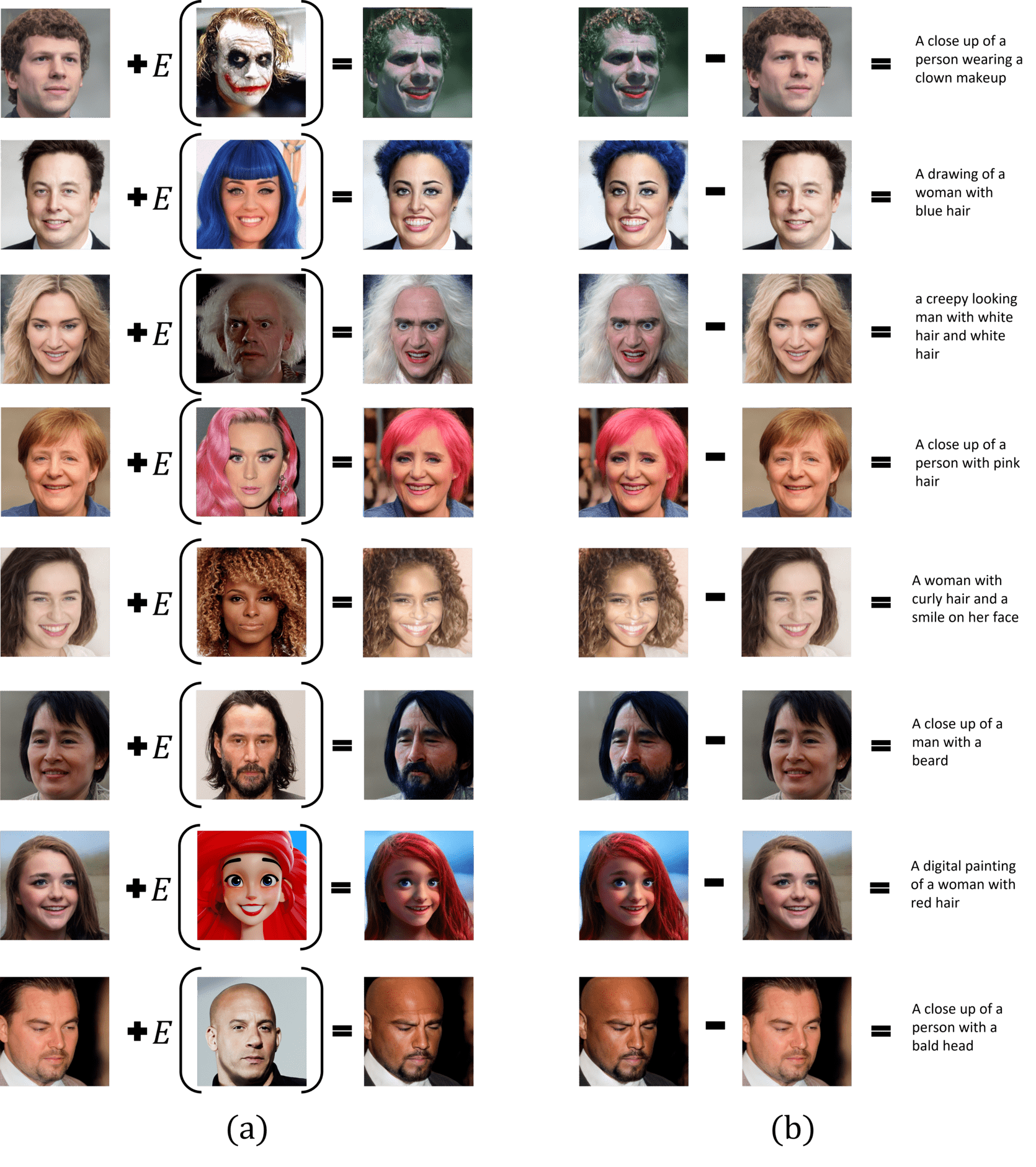} 
    \captionof{figure}{Examples of essence decoding. (a) presents the targets, sources, and manipulation results, with $E$ representing the essence extraction, i.e. we add the essence of the right image to the left image, and (b) demonstrates the decoding of the essence vectors for each example. 
    }
    \label{fig:interpret_supp}
\end{center}
    \end{figure}

\
Fig.~\ref{fig:interpret_supp} presents additional examples of essence decoding with BLIP for various sources and targets using our optimizer. The decoding demonstrates that the transferred attributes correspond to the most notable semantic attributes of the target images.

\clearpage
\section{Ablation Study}

\begin{figure}[h]
    \begin{tabular*}{\linewidth}{@{\extracolsep{\fill}}lcc@{~~}c@{~~}c@{~~}c@{~~}c@{~~}c}
        \toprule
         &  & Quality &\multicolumn{2}{c}{Identity scores} & \multicolumn{2}{c}{Semantic scores}\\
        \cmidrule(lr){4-5}
        \cmidrule(lr){6-7}
        & &  FID ($\downarrow$) & Source ($\uparrow$)& Target ($\downarrow$) & BLIP ($\uparrow$) & CLIP ($\uparrow$)\\
        \midrule
        &\textbf{Ours} & 173.4$\pm$24.6 & 42.6$\pm$11.1 & 17.4$\pm$8.1 & 67.5$\pm$6.7 & 73.5$\pm$5.1 \\
        &\textbf{Our w/o Eq. 6} & 165.6$\pm$19.5 & 45.6$\pm$9.8 & 15.5$\pm$6.6 & 63.9$\pm$7.8  & 69.5$\pm$4.1 \\
        &\textbf{Our w/o Eq. 5}& \textbf{133.0$\pm$0} &  \textbf{77.4$\pm$0} & \textbf{2.2$\pm$5.0} & {\color{Mahogany}51.2$\pm$6.6} & {\color{Mahogany}53.2$\pm$3.4} \\
        &\textbf{Our w/o $L_2$} & {\color{orange}233.3$\pm$29.9} & {\color{orange}6.8$\pm$3.8} & 26.6$\pm$8.9 & \textbf{78.2$\pm$6.2} & \textbf{88.3$\pm$4.0} \\
        \bottomrule
    \end{tabular*}
    \captionof{table}{Quantitative comparison of different variations of our optimization-based method on the celebrities test.
    Results that indicate identity loss of the source are marked in orange; results that indicate that no semantic attributes were transferred are marked in red. 
    }
    \label{tab:ablation}
\end{figure}

Our ablation tests are conducted with the optimizer, which is also the most successful variant of our method, using the celebrity targets set. We choose to use the celebrity targets dataset due to its diverse nature and challenging semantic attributes. We use the first $4$ images of StyleCLIP's test set as our training sources.

As can be seen from Fig.~8 of the main text, removing the similarity loss (Eq.~5) results in nearly no semantic change to the sources. As explained, this can be attributed to the fact that Eq.~5 is the only one among our loss terms that demands semantic similarity to the target. When removing the $L_2$ regularization, the result is a blurry unified identity presenting some of the semantic attributes of the target. Fig.~8 of the main text also demonstrates that our consistency loss (Eq.~6) is necessary to produce semantically accurate directions, otherwise, the optimization deviates towards a partial subset of the semantic properties that are easy to control for the source images.  

\begin{figure}[h]
    \begin{center}
    \renewcommand{\arraystretch}{0.2}
\begin{tabular}{@{}c@{}c@{~}c@{~}c@{~}c@{~~~}c@{~}c@{~}c@{~~~}c@{~}c@{~}c@{~~~}c@{~}c@{~}c@{}}
\centering{  Target }
&
\centering{\begin{turn}{90}~ Src. \end{turn} }
&
\includegraphics[width=0.065\linewidth,clip] {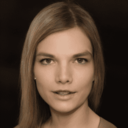}&
\includegraphics[width=0.065\linewidth,clip] {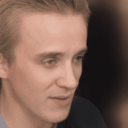}&
\includegraphics[width=0.065\linewidth,clip] {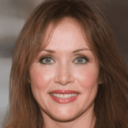}&
\includegraphics[width=0.065\linewidth,clip] {ablations_supp/1.png}&
\includegraphics[width=0.065\linewidth,clip] {ablations_supp/2.png}&
\includegraphics[width=0.065\linewidth,clip] {ablations_supp/3.png}&
\includegraphics[width=0.065\linewidth,clip] {ablations_supp/1.png}&
\includegraphics[width=0.065\linewidth,clip] {ablations_supp/2.png}&
\includegraphics[width=0.065\linewidth,clip] {ablations_supp/3.png}&
\includegraphics[width=0.065\linewidth,clip] {ablations_supp/1.png}&
\includegraphics[width=0.065\linewidth,clip] {ablations_supp/2.png}&
\includegraphics[width=0.065\linewidth,clip] {ablations_supp/3.png}
\\
\includegraphics[width=0.065\linewidth, clip]{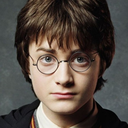}
&
&
\includegraphics[width=0.065\linewidth,clip] {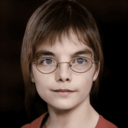} 
&
\includegraphics[width=0.065\linewidth,clip] {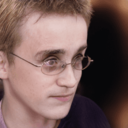} &
\includegraphics[width=0.065\linewidth, clip]{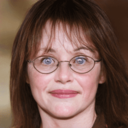}  
&
\includegraphics[width=0.065\linewidth,clip] {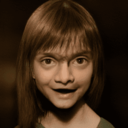} 
&
\includegraphics[width=0.065\linewidth,clip] {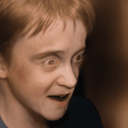} 
&
\includegraphics[width=0.065\linewidth,clip] {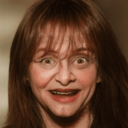} &
\includegraphics[width=0.065\linewidth, clip]{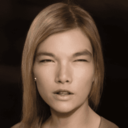}  
&
\includegraphics[width=0.065\linewidth,clip] {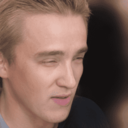} 
&
\includegraphics[width=0.065\linewidth,clip] {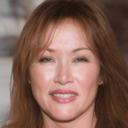} 
&
\includegraphics[width=0.065\linewidth,clip] {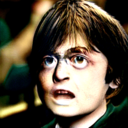} &
\includegraphics[width=0.065\linewidth, clip]{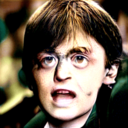}  
&
\includegraphics[width=0.065\linewidth,clip] {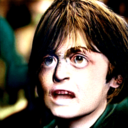}
\\
\includegraphics[width=0.065\linewidth, clip]{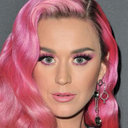}
&
&
\includegraphics[width=0.065\linewidth,clip] {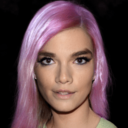} 
&
\includegraphics[width=0.065\linewidth,clip] {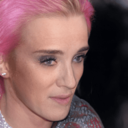} &
\includegraphics[width=0.065\linewidth, clip]{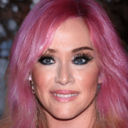}  
&
\includegraphics[width=0.065\linewidth,clip] {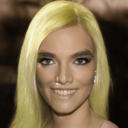} 
&
\includegraphics[width=0.065\linewidth,clip] {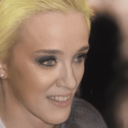} 
&
\includegraphics[width=0.065\linewidth,clip] {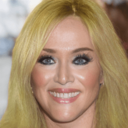} &
\includegraphics[width=0.065\linewidth, clip]{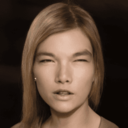}  
&
\includegraphics[width=0.065\linewidth,clip] {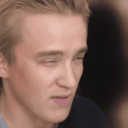} 
&
\includegraphics[width=0.065\linewidth,clip] {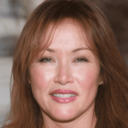} 
&
\includegraphics[width=0.065\linewidth,clip] {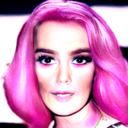} &
\includegraphics[width=0.065\linewidth, clip]{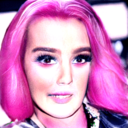}  
&
\includegraphics[width=0.065\linewidth,clip] {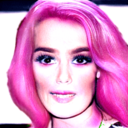}
\\
\includegraphics[width=0.065\linewidth, clip]{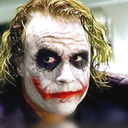}
&
&
\includegraphics[width=0.065\linewidth,clip] {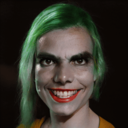} 
&
\includegraphics[width=0.065\linewidth,clip] {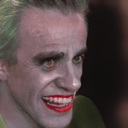} &
\includegraphics[width=0.065\linewidth, clip]{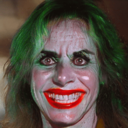}  
&
\includegraphics[width=0.065\linewidth,clip] {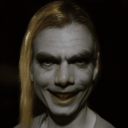} 
&
\includegraphics[width=0.065\linewidth,clip] {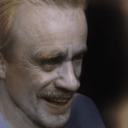} 
&
\includegraphics[width=0.065\linewidth,clip] {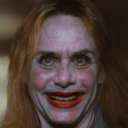} &
\includegraphics[width=0.065\linewidth, clip]{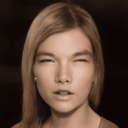}  
&
\includegraphics[width=0.065\linewidth,clip] {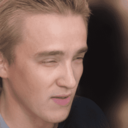} 
&
\includegraphics[width=0.065\linewidth,clip] {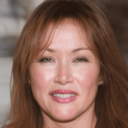} 
&
\includegraphics[width=0.065\linewidth,clip] {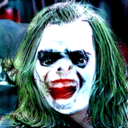} &
\includegraphics[width=0.065\linewidth, clip]{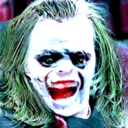}  
&
\includegraphics[width=0.065\linewidth,clip] {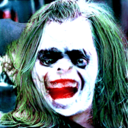}
\\
\includegraphics[width=0.065\linewidth, clip]{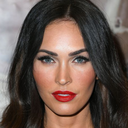}
&
&
\includegraphics[width=0.065\linewidth,clip] {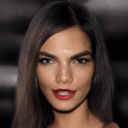} 
&
\includegraphics[width=0.065\linewidth,clip] {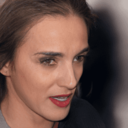} &
\includegraphics[width=0.065\linewidth, clip]{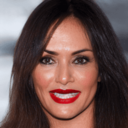}  
&
\includegraphics[width=0.065\linewidth,clip] {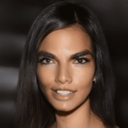} 
&
\includegraphics[width=0.065\linewidth,clip] {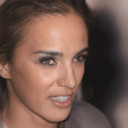} 
&
\includegraphics[width=0.065\linewidth,clip] {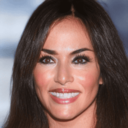} &
\includegraphics[width=0.065\linewidth, clip]{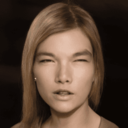}  
&
\includegraphics[width=0.065\linewidth,clip] {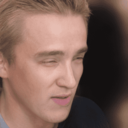} 
&
\includegraphics[width=0.065\linewidth,clip] {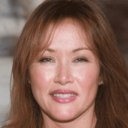} 
&
\includegraphics[width=0.065\linewidth,clip] {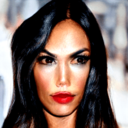} &
\includegraphics[width=0.065\linewidth, clip]{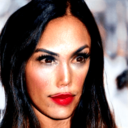}  
&
\includegraphics[width=0.065\linewidth,clip] {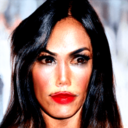}
\\
\includegraphics[width=0.065\linewidth, clip]{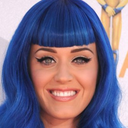}
&
&
\includegraphics[width=0.065\linewidth,clip] {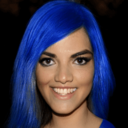} 
&
\includegraphics[width=0.065\linewidth,clip] {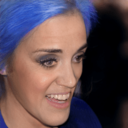} &
\includegraphics[width=0.065\linewidth, clip]{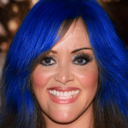}  
&
\includegraphics[width=0.065\linewidth,clip] {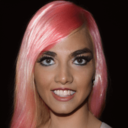} 
&
\includegraphics[width=0.065\linewidth,clip] {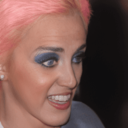} 
&
\includegraphics[width=0.065\linewidth,clip] {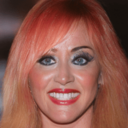} &
\includegraphics[width=0.065\linewidth, clip]{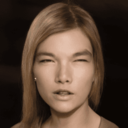}  
&
\includegraphics[width=0.065\linewidth,clip] {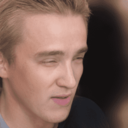} 
&
\includegraphics[width=0.065\linewidth,clip] {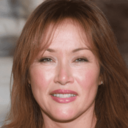} 
&
\includegraphics[width=0.065\linewidth,clip] {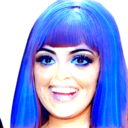} &
\includegraphics[width=0.065\linewidth, clip]{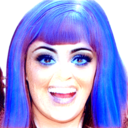}  
&
\includegraphics[width=0.065\linewidth,clip] {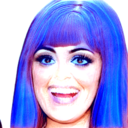}
\\
\includegraphics[width=0.065\linewidth, clip]{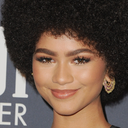}
&
&
\includegraphics[width=0.065\linewidth,clip] {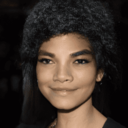} 
&
\includegraphics[width=0.065\linewidth,clip] {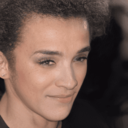} &
\includegraphics[width=0.065\linewidth, clip]{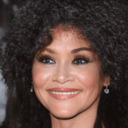}  
&
\includegraphics[width=0.065\linewidth,clip] {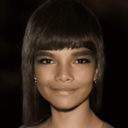} 
&
\includegraphics[width=0.065\linewidth,clip] {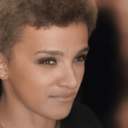} 
&
\includegraphics[width=0.065\linewidth,clip] {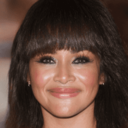} &
\includegraphics[width=0.065\linewidth, clip]{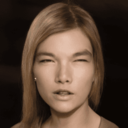}  
&
\includegraphics[width=0.065\linewidth,clip] {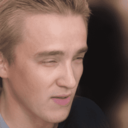} 
&
\includegraphics[width=0.065\linewidth,clip] {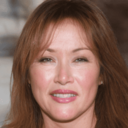} 
&
\includegraphics[width=0.065\linewidth,clip] {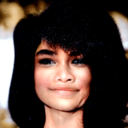} &
\includegraphics[width=0.065\linewidth, clip]{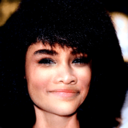} 
&
\includegraphics[width=0.065\linewidth, clip]{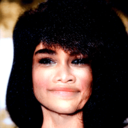} 
\\\\
\cmidrule(r){3-5}
\cmidrule(r){6-8}
\cmidrule(r){9-11}
\cmidrule(r){12-14}
 & & & \multicolumn{1}{c}{Ours} & &   \multicolumn{3}{c}{Ours w/o Eq.~6}&  \multicolumn{3}{c}{Ours w/o Eq.~5} & \multicolumn{3}{c}{Ours w/o $L_2$}\\
\end{tabular}
    \captionof{figure}{A comparison between different variations of our method. 
    Removing the consistency loss (Eq.~6) results in inaccurate or partial transfer of the attributes, removing the similarity loss (Eq.~5) results in nearly no semantic changes to the source, and removing the $L_2$ regularization results in severe identity loss.
    }
    \label{fig:ablations_supp}
\end{center}
    \end{figure}

Tab.~\ref{tab:ablation} encloses the results of our full ablation tests. As can be seen, removing our $L_2$ loss results in severe identity loss, with the overall lowest source identity score by a very large margin. This result is consistent with Fig.~8 in the main text. Removing Eq.~5 results in very low semantic scores and great identity preservation, as can be expected since there is almost no change in the source image. Note that in the case where we remove Eq.~5, the algorithm is target-agnostic, therefore the FID and source identity scores have a standard deviation of $0$. When removing Eq.~6, notice how both semantic scores are degraded. This can be attributed to the fact that as mentioned, the consistency loss is necessary to ensure proper transfer of the semantic attributes, especially when the target presents challenging attributes such as unusual hair colors or styles, glasses, and makeup (see Fig.~\ref{fig:ablations_supp}). While the identity scores are slightly improved without Eq.~6, recall that as mentioned in the main text, there is an inherent trade-off between the identity scores and the semantic scores- as the semantic scores increase, more attributes are transferred, thus the identity preservation score is lower as a direct result. Importantly, the identity preservation property from the main text $\text{ID-score}_{source} > \text{ID-score}_{target}$ is well preserved with our method, and the target identity score is very similar and low in both cases (with and without Eq.~6), indicating that the added semantic changes by the consistency do not cause the identity to shift more towards the target than the source, i.e. our consistency loss in Eq.~6 is necessary to transfer the essence features that \emph{do not} break the balance between the source and target identities, and maintains high quality, diverse results, as reflected by the higher semantic scores. 

Fig.~\ref{fig:ablations_supp} contains additional examples from the described ablation experiment. As can be seen, in accordance with Tab.~\ref{tab:ablation}, removing Eq.~6 results in inaccurate or partial semantic transfer. For example, with Harry Potter as target (first row in Fig.~\ref{fig:ablations_supp}), the signature glasses are not transferred without our consistency loss (without Eq.~6), while our full method captures the glasses in their original shape, as well as the hair color, and the pale skin. In other cases, removing Eq.~6 leads to inaccurate transfer, for example, without Eq.~6, the pink and blue hair (lines 2,5 of Fig.~\ref{fig:ablations_supp}) transfer blonde-green and orange-pink hair, respectively, while our full method faithfully captures all hair colors. Removing Eq.~5 leads to nearly no semantic change as expected, and removing the $L_2$ regularization results in severe identity loss and distorted results. 

\clearpage
\section{Sensitivity Test}
\begin{figure*}[h]%
    \centering
    \begin{tabular}{c@{~~~~~~~~~~~~~~~~}c@{}}
    {{\includegraphics[width=5cm]{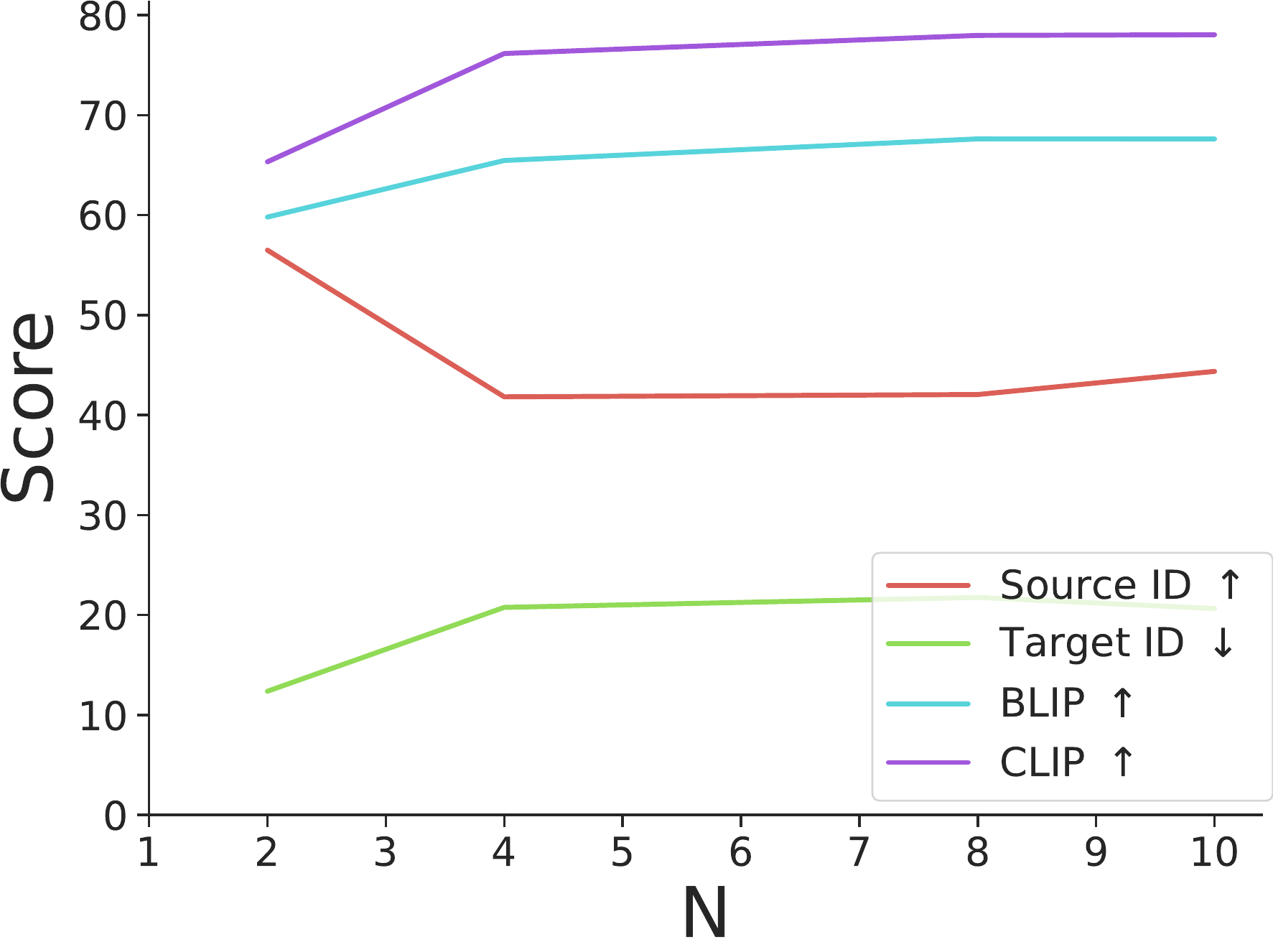} }}&
    {{\includegraphics[width=5cm]{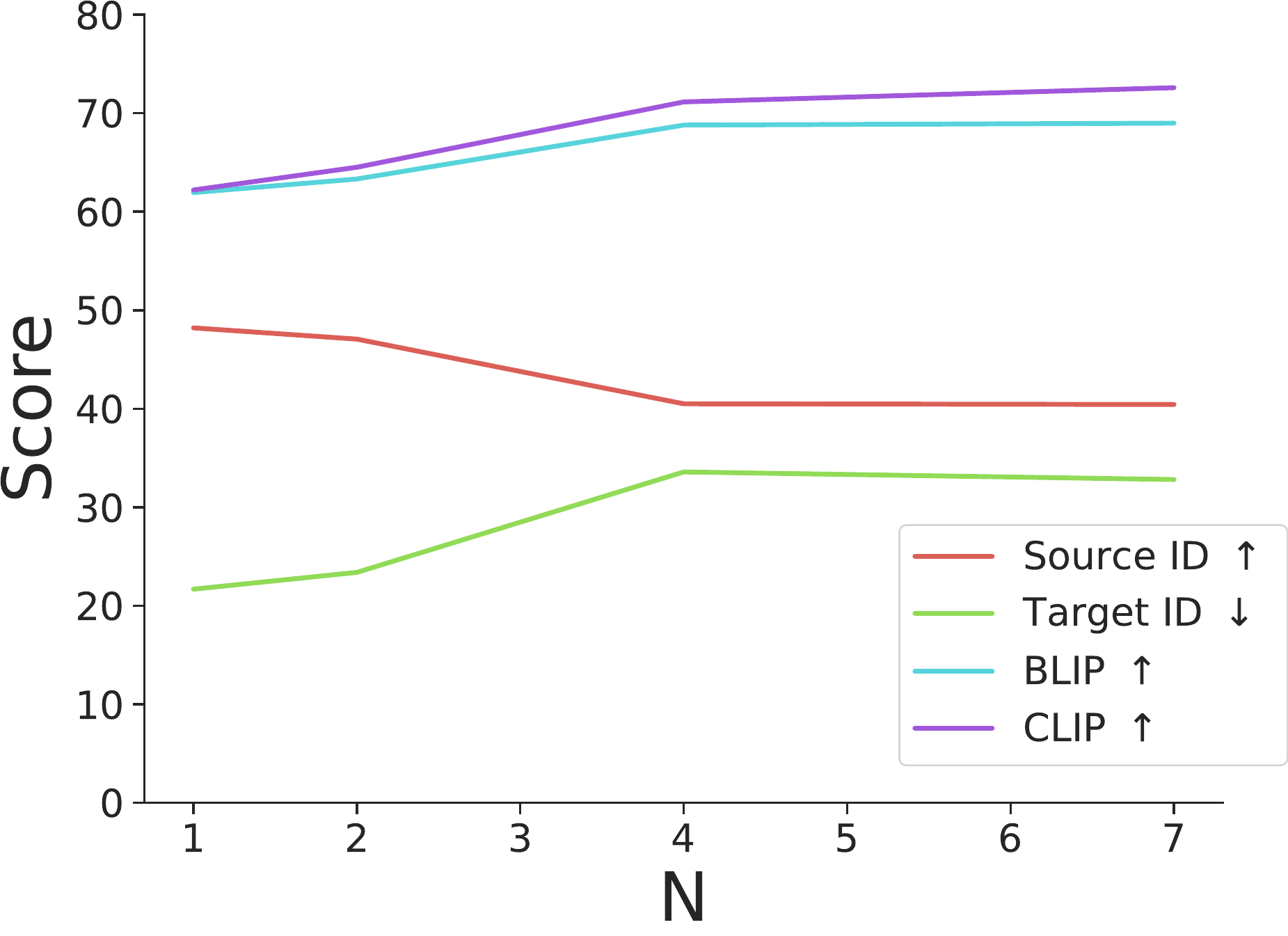}}}
    \\
    (a) & (b)\\
    \end{tabular}
    \caption{Sensitivity tests to the number of training sources $N$ with (a) the optimizer, (b) the encoder.}%
    \label{fig:sensitivity}%
\end{figure*}

To evaluate our method's sensitivity to the selection of the number of training sources $N$, we conduct an experiment where we gradually increase $N$ on our optimizer and encoder, using random targets from the celebrities test. As can be seen from Fig.~\ref{fig:sensitivity}, for small values of $N$ ($N<4$) the semantic BLIP and CLIP scores are low indicating that the semantic features of the target were not transferred to the sources. Starting from $N=4$, the semantic scores are high, at the expanse of some identity loss. There is some evident advantage to using more than $N=4$ sources for training, however, a larger training set would require additional computational resources for training. Additionally, in accordance with the results in the main paper, the optimizer is superior to the encoder in terms of identity preservation (low target ID score, even for large values of $N$).

\end{document}